\documentclass[lettersize,journal]{IEEEtran}
\usepackage{amsmath,amsfonts}

\usepackage{algorithm}
\usepackage{algpseudocode}
\usepackage{array}
\usepackage[caption=false,font=normalsize,labelfont=sf,textfont=sf]{subfig}
\usepackage{textcomp}
\usepackage{stfloats}
\usepackage{url}
\usepackage{verbatim}
\usepackage{booktabs}
\usepackage{graphicx}
\usepackage{makecell}
\usepackage[table,xcdraw]{xcolor}
\usepackage{cite}
\usepackage{soul}
\usepackage{bm}
\usepackage{amssymb}

\usepackage[pagewise]{lineno}
\usepackage{etoolbox}

\usepackage{ulem}     
\usepackage[hidelinks]{hyperref}

\makeatletter

\newcommand{\Rmnum}[1]{\expandafter\@slowromancap\romannumeral #1@}
\makeatother

\begin{document}


\title{Unsupervised Cross-Domain 3D Human Pose Estimation via Pseudo-Label-Guided Global Transforms}

\author{Jingjing Liu, Zhiyong Wang, Xinyu Fan, Amirhossein Dadashzadeh, Honghai Liu,~\IEEEmembership{Fellow,~IEEE}, Majid Mirmehdi

\thanks{This work was supported by the TORUS Project, which has been funded by the UK Engineering and Physical Sciences Research Council (EPSRC), grant number EP/X036146/1.}
\thanks{J. Liu (Corresponding author), A. Dadashzadeh and M. Mirmehdi are with the School of Computer Science, University of Bristol, United Kingdom. (e-mail: jingjing.liu@bristol.ac.uk)}
\thanks{Z. Wang and H. Liu are with the State Key Laboratory of Robotics and Systems, Harbin Institute of Technology Shenzhen, China.}
\thanks{Xinyu Fan is with the School of Aerospace Engineering, Xiamen University, China.}
}


\IEEEpubid{\begin{minipage}{\textwidth}\ \\[12pt]
\centering
\footnotesize
Copyright~\copyright~2025 IEEE. Personal use of this material is permitted.
However, permission to use this material for any other purposes must be obtained
from the IEEE by sending an email to pubs-permissions@ieee.org.
\end{minipage}}

\maketitle

\begin{abstract}

Existing 3D human pose estimation methods often suffer in performance, when applied to cross-scenario inference, due to domain shifts in characteristics such as camera viewpoint, position, posture, and body size. Among these factors,  camera viewpoints and locations {have been shown} to contribute significantly to the domain gap by influencing the global positions of human poses.
To address this, we propose a novel framework that explicitly conducts global transformations between pose positions in the camera coordinate systems of source and target domains.
We start with a Pseudo-Label Generation Module that is applied to the 2D poses of the target dataset to generate pseudo-3D poses. Then, a Global Transformation Module leverages a human-centered coordinate system as a novel bridging mechanism to seamlessly align the positional orientations of poses across disparate domains, ensuring consistent spatial referencing.
To further enhance generalization, a Pose Augmentor is incorporated to address variations in human posture and body size. This process is iterative, allowing refined pseudo-labels to progressively improve guidance for domain adaptation.
Our method is evaluated on various cross-dataset benchmarks, including Human3.6M, MPI-INF-3DHP, and 3DPW. The proposed method outperforms state-of-the-art
approaches and even outperforms the target-trained model.
\end{abstract}


\section{Introduction}

\IEEEPARstart{H}{uman} pose estimation has a wide range of applications, including video surveillance \cite{jiao2020pen}\cite{wu2023video}, augmented reality \cite{guzov2021human, yi2023mime, haouchine2021pose}, motion analysis \cite{yu2023toward}\cite{lu2023hard}, robotics \cite{zimmermann20183d}\cite{liu2021collision}, and human-computer interaction \cite{svenstrup2009pose}\cite{huo20233d}.
While current methods for 3D human pose estimation have made significant progress, acquiring large amounts of diverse 3D data remains a major challenge. Most existing approaches \cite{gu2019multi,wandt2021canonpose,tang2023ftcm,zhou2023dual} rely on datasets collected in controlled laboratory environments, while the variability of data is often limited.
As a result, when these models are applied to other significantly different scenarios, their performances tend to degrade due to the lack of diversity in the training data. This highlights the need to tackle the challenges of achieving adaptability in different scenarios.

\begin{figure}[t]
    \centerline{\includegraphics[width=8cm]{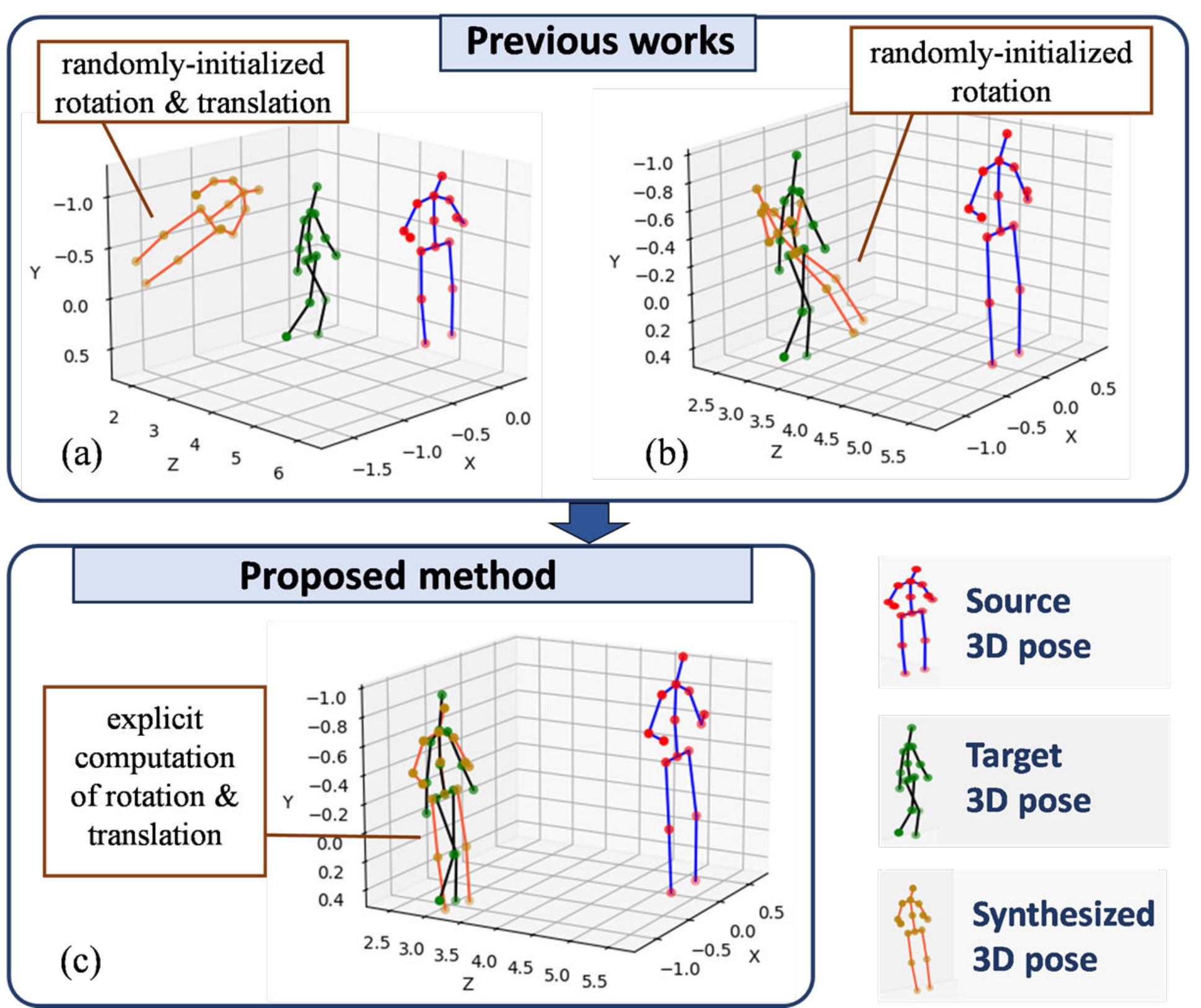}}
    \caption{{\bf Addressing the domain gap over camera viewpoints and locations.} {Rigid transformation operations (rotation and translation) are used to control pose adaptation to variations in camera viewpoints and positions.}
    Most previous works synthesized data from source 3D poses through (a) randomly initialized rotation and translation {within adversarial learning} and (b) explicit computation of the translation matrices but still relied on randomly initialized rotation matrices {within adversarial learning}. In our method, (c) we propose explicit computation of both the rotation and translation matrices to align the source and target data.
    }
    \label{fig-overview} 
\end{figure}

The domain gap in 3D human pose estimation typically arises from factors such as camera viewpoints, human postures, actions, and even body sizes \cite{zhang2020inference,gong2021poseaug,gholami2022adaptpose,peng2023source}. To address such shortcomings, some works have focussed on the task of unsupervised cross-domain 3D human pose estimation \cite{gong2021poseaug,gholami2022adaptpose,peng2023source,peng2024dual}. In this setting, paired 2D and annotated 3D poses from the source domain are provided, while the target domain only offers 2D poses without corresponding 3D annotations. The objective is to train a model using all this data and then apply the model to infer 3D human poses in the target domain.

\IEEEpubidadjcol
A pioneering work is \cite{gong2021poseaug}, which proposed to augment the available training poses via adversarial training. {Their augmentor learnt to} change various factors, such as the bone angle, bone length, and rigid transformation of the whole pose (rotation and translation) to create augmented data. Their 3D pose estimator was better trained to generalize to other domains, given such more diverse data. In \cite{peng2024dual}, two pose augmentors, combined with meta-optimization, were used to explore out-of-source distributions. This is particularly significant in cases where target poses differ significantly from source distributions. Applying a weak augmentor as an intermediary, a strong augmentor was designed to generate poses that exhibit significant deviations from the source distributions. Both \cite{gong2021poseaug} and \cite{peng2024dual} belong to Domain Generalization (DG) methods, in which only the source data is incorporated for the cross-domain learning.

Domain Adaptation (DA) is another type of approach for cross-domain learning by utilizing the available target domain data (i.e., input 2D poses) to improve cross-domain performance. In \cite{gholami2022adaptpose}, target 2D poses were used to enforce their generated synthetic data to have similar distributions. The data augmentation process in \cite{gong2021poseaug,gholami2022adaptpose,peng2024dual} {relied on randomly initialized GAN-based adversarial training that only captures the source data distributions (e.g. see Fig. \ref{fig-overview} (a)). Since the synthesized data are randomly augmented, these methods are not efficient when adapting to a different domain.}

Some other works have considered the knowledge embedded in the target dataset to effectively generate poses that simulate the target domain.
For example, to address the large distribution gap due to camera positions, \cite{chai2023global} exploited the direct use of the camera's intrinsic parameters to compute the transformation between the source and target poses. The translation matrices were explicitly calculated, while the rotations were obtained through a GAN-based augmentor that also considered bone vector and bone length {(see Fig. \ref{fig-overview} (b)).}  In \cite{liu2023posynda},  a diffusion-based model was applied to generate multiple 3D pose hypotheses using target 2D poses for conditioning, thus simulating the target distribution. The generated 3D pose hypotheses were then directly used as pseudo-labels to train the pose estimator. However, without proper discriminators to ensure their plausibility, the generated poses may include anatomically implausible joint configurations such as unrealistic joint angles.

In this paper, we address the domain discrepancy in 3D human pose estimation, predicated on camera viewpoints and positions.
Motivated by recent methods that exploit information embedded in the target dataset, we proposed to align the 3D poses between the source and target camera coordinate systems through a target-domain-driven global transformation.
Here, the alignment refers to transforming the source poses through rotation and translation to minimize the average distance between the transformed source poses and the target poses over corresponding joints. In this way, the discrepancy in terms of their 3D absolute coordinates is mitigated. Additionally, the global transformation refers to the transformation between absolute coordinates under camera coordinate systems rather than root-relative coordinates {such  that 2D poses can also be aligned after camera projection.}
Compared with augmenting rotation and translation matrices through GAN-based adversarial learning in which the parameters are randomly initialized, our target-driven global transformation is more intuitive and effective in adapting to camera viewpoints and positions (see Fig. \ref{fig-overview}(c)). The explicit computation of rotation and translation matrices makes the synthesized data align with the target data in space as closely as possible.

Given the 2D poses of the target dataset, we propose a Pseudo-Label Generation Module to generate 3D poses as pseudo-labels. Based on the root-relative predictions from pre-trained models  \cite{pavllo20193d, zhang2022mixste}, we leverage bone length consistency and 2D reprojection constraints to infer absolute 3D joint positions as the pseudo-labels. In contrast to most prior works \cite{gong2021poseaug,gholami2022adaptpose, chai2023global, peng2023source, peng2024dual} which overlook this aspect, we explicitly explore both the generation and effective use of pseudo-labels of the target domain. In our framework, these pseudo-labels play a critical role in guiding the synthesis of data to ensure alignment with the target domain.
In order to address variations in human posture and body size, we incorporate a GAN-based Pose Augmentor which generates sequences of 3D human poses by changing bone angles and lengths.
Next, to align 3D poses between the augmented source domain and the target domain, we introduce a novel human body-centered coordinate system, which serves as a bridge in the rotation and translation process. Then the aligned 3D poses are projected into 2D using the camera intrinsic parameters of the target domain, resulting in aligned 2D poses as well. These sequences and corresponding re-projected 2D pose sequences are subsequently fed into two discriminators for better fidelity.
{These synthesized 2D-3D data pairs are also used to fine-tune a pre-trained 3D pose estimation model. At each iteration, the pose estimator generates increasingly more accurate pseudo-labels to better adapt to the target domain.}
The above modules are integrated into a unified framework, cohesively bridging the domain gap and substantially improving 3D pose estimation performance when adapted to the target domain.

Our main contributions can be summarized as follows: (i) We propose a novel global transformation method to reduce the domain discrepancies caused by varying camera parameters. This approach is significantly resilient to camera position and viewpoint variations, without introducing any additional training parameters. (ii)
We exploit the target dataset's latent knowledge and generate 3D pseudo-labels that better approximate the target distribution to guide the global transformation. (iii) We demonstrate the effectiveness and efficiency of the proposed method through experiments and ablations on three benchmarks, Human3.6M {(H3.6M)}, MPI-INF-3DHP (3DHP), 3DPW and achieve state-of-the-art results on {H3.6M-3DHP and H3.6M-3DPW} cross-dataset evaluation, outperforming the lastest state-of-the-art method by 7.2\% and 11.8\% respectively. Our cross-dataset evaluation results in an unsupervised setting on the target dataset even surpass the target-specific trained MixSTE model \cite{zhang2022mixste}.
Additionally, we simulate a new viewpoint based on the 3DHP dataset to further validate the robustness of our method.

\section{Related Work}

\subsection{3D Human Pose Estimation}

Recent years have seen significant advances in skeleton-based 3D human pose estimation, which can be categorized into two primary approaches: (a) One-stage methods, such as \cite{lin2021end}, \cite{jin2022single}, \cite{li2023niki}, and  \cite{sarandi2023learning}, which directly predict 3D poses from input images or videos; (b) Two-stage methods, exemplified by \cite{zhang2022mixste,zheng20213d,zou2021modulated,li2022mhformer,tang20233d}, which first estimate 2D poses from images or videos \cite{sun2019deep,li2020cascaded,jin2020whole} and subsequently lift these to 3D.
Martinez et al. \cite{martinez2017simple} conducted a pioneering work demonstrating that lifting ground truth 2D joint positions to 3D can be accomplished with very low errors. As 2D pose estimation techniques are quite mature \cite{fang2022alphapose, lu2024rtmo, liu2024human}, we only focus on recent methods that lift 2D poses to 3D.

In \cite{zou2021modulated}, a {novel Graph Convolutional Network} was proposed for 3D HPE. Weight modulation and affinity modulation were designed to assign unique modulation vectors to nodes and update the graph structure respectively, thereby improving performance without increasing model size. With the rise of transformers,
PoseFormer \cite{zheng20213d} was the first purely transformer-based 3D pose estimation approach in which a spatial transformer encoded the local relationships between the 2D joints and a temporal transformer captured global dependencies across the arbitrary frames. Later, Zhang et al. \cite{zhang2022mixste} proposed MixSTE, which took a sequence of 2D keypoints as input and predicted 3D poses. It alternately applied spatial and temporal transformer blocks to capture spatio-temporal dependencies. Li et al. \cite{li2022mhformer} introduced MHFormer, which generated multiple pose hypotheses via a multi-hypothesis transformer to better model temporal uncertainty. {Tang} et al. \cite{tang20233d} further proposed a two-pathway block that modelled spatial and temporal information in parallel.

More recently, diffusion models have been explored to generate multiple hypotheses for 3D pose estimation, e.g.  \cite{shan2023diffusion}\cite{cai2024}. These often involve a learned reverse process where noise is removed step-by-step to recover the true pose. Shan et al. \cite{shan2023diffusion} proposed a joint-level aggregation method based on reprojection errors from multiple hypotheses and achieved exceptional performance. Rather than directly regressing the 3D joint coordinates as in \cite{shan2023diffusion}, the work in \cite{cai2024} decomposed the 3D pose into bone length and direction in the forward diffusion process to speed up gradient descent. Although these methods perform well on their training dataset, their performance tends to degrade when generalized to other datasets \cite{liu2023posynda,peng2024dual}.

\subsection{Data augmentation on 3D human poses}
To improve the generalization ability of a model,
previous approaches used data augmentation to increase the diversity of training data \cite{rogez2016mocap, mehta2017vnect, xu2021monocular, yang2018body, gong2021poseaug, gholami2022adaptpose, yang2023camerapose, du2024joypose,  peng2024dual}.
Some methods \cite{rogez2016mocap, mehta2017vnect,yang2018body} focused on augmenting the original images, but more relevant to our method are those that enhanced 2D-3D pose pairs \cite{gong2021poseaug, gholami2022adaptpose, yang2023camerapose, du2024joypose, peng2024dual, kimtoward}.
{In the latter methods, the data augmentor is jointly trained with the discriminator and the pose estimator end-to-end with an error-feedback training strategy. As such, the augmentor learns to augment data with guidance from the estimator and discriminator in an unsupervised way (also see Section \ref{sec:unsup-review}). The augmented data are used to train the pose estimator, thus improving the generalization and adaptation ability of the pose estimator.}
For example, in PoseAug \cite{gong2021poseaug}, data augmentation was applied in terms of bone angles, lengths, and body rigid transformations. More recently, Peng et al. \cite{peng2024dual} proposed a dual-augmentor framework, which included both a weak and a strong pose augmentor. This approach preserved knowledge of source poses while exploring out-of-source distributions. \cite{kimtoward} proposed a Biomechanical Pose Generator (BPG), a novel approach that harnessed biomechanical principles, particularly the normal range of motion, to autonomously generate diverse and realistic 3D poses. This approach did not rely on the source dataset, thus overcoming the restrictions of popularity bias.

\begin{figure*}[]
    \centerline{\includegraphics[width=17cm]{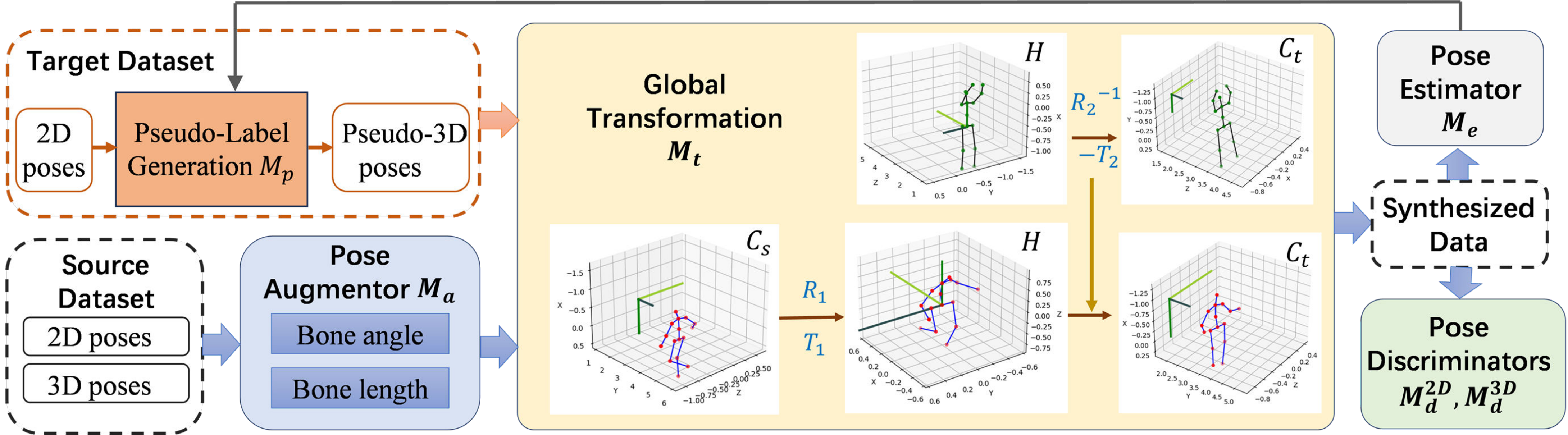}}
    \caption{{\bf Overview of the proposed method.} The pseudo-3D poses from target dataset are used to guide the global transformation of the augmented 3D poses from source dataset. During the transformation, a human-centric coordinate system is proposed as a bridge. The synthesized 2D-3D pose pairs are then used to train the discriminators $M_d^{2D}$, $M_d^{3D}$ and Pose Estimator $M_e$ jointly. The fine-tuned Pose Estimator helps generate more accurate pseudo-3D poses of the target dataset, making the framework iterative. In the Global Transformation module $M_t$, the symbols $C_s$ and $C_t$ represent the camera coordinate systems for the source ($s$) and target ($t$) domains, respectively. $H$ denotes the human-centric coordinate system. $R_1$, $R_2$ represent the rotation matrices among coordinate systems and $T_1$, $T_2$ represent the translation matrices.}
    \label{fig-method}
\end{figure*}

\subsection{Unsupervised Domain Adaptation in 3D Pose Estimation} \label{sec:unsup-review}
Unsupervised domain adaptation in 3D human pose estimation involves transferring models from the fully labelled source domain to an unlabelled target domain, with several different strategies seen in recent works. Zhang et al. \cite{zhang2020inference} proposed an Inference Stage Optimization which performed geometry-aware self-supervised learning (SSL) on each single target instance and updated the 3D pose model before making predictions. Guan et al. \cite{guan2021bilevel} used 2D information of individual test frames and temporal constraints to fine-tune the source model on test video streams.

Recently, some works have aimed to further improve performance by synthesizing data from the source dataset in other ways \cite{chai2023global, liu2023posynda}. The synthesized data are typically designed to match the distribution of the target domain, and then used to train the pose estimator, to improve its adaptation ability.
For example, to address the domain gap in camera viewpoints, Chai et al. \cite{chai2023global} exploited geometric constraints to align the root joint positions of 3D poses between the source and target domains. While the alignment of the root joint was achieved through the explicit computation of the translation matrix, the alignment of the remaining joints relied on the rotation matrix, which was still learnt via GAN-based methods.  In \cite{liu2023posynda}, the diffusion-based human pose estimator D3DP \cite{shan2023diffusion} was deployed to generate pseudo 2D-3D pairs of the target domain, which were directly used to train the pose estimator. However, without checking for plausibility, their generated 3D pose hypotheses may exhibit unreasonable joint angles. The above methods attempted to simulate the target domain via exploration of the latent knowledge within the target domain and achieved some improvements. Inspired by them, we aim to more fully exploit target domain information by generating pseudo-labels to guide synthetic data while ensuring their plausibility.

Next, we introduce a method that proposes global transformations to address the variations in camera viewpoints and positions, which are key contributors to the domain gap \cite{gong2021poseaug}\cite{gholami2022adaptpose}.
Guided by the target domain, our approach aims to align both 2D and 3D poses between the source and target domains to match the target data distribution as closely as possible. The proposed method integrates a Global Transformation module with a Pose Augmentor that augments bone angles and lengths. This combination achieves state-of-the-art performance in cross-domain 3D human pose estimation tasks.

\section{Method}
\label{sec:method}

\subsection{Problem Formulation}
\label{sec-prob}

{Unsupervised cross-domain 3D human pose estimation aims to adapt a pose estimation model trained on a labelled source dataset to an unlabelled target dataset.}
Let the human pose representation consist of $J$ skeleton joints, numbered from $0$ to $J-1$. We define $\bm{P}_{2D}^{src}\in R^{2\times J}$  to denote the 2D pose and $\bm{P}_{3D}^{src}\in R^{3\times J}$ to denote the label, i.e., a 3D pose from the source dataset and,
$\bm{P}_{2D}^{tgt}\in R^{2\times J}$ to represent a 2D pose from the target dataset which does not have any 3D pose label.  
Given the intrinsic parameters $\bm{K}$ of the camera from the target dataset, our method synthesizes augmented and aligned 
2D and 3D pose pairs $\bm{P}^{'} =(\bm{P}_{2D}^{'}, \bm{P}_{3D}^{'})$ with 
\begin{equation}
\begin{array}{l}
\bm{P}_{3D}^{'}=G(\bm{P}_{2D}^{src},\bm{P}_{3D}^{src},\bm{P}_{2D}^{tgt}) , \\
 \bm{P}_{2D}^{'}=F(\bm{P}_{3D}^{'},\bm{K}) ,
  \end{array}
\label{eq1}
\end{equation}
where $G$ denotes a generator function and $F$ is the projection function from 3D camera coordinates to 2D image coordinates.
The synthesized pose pairs are used to fine-tune the Pose Estimator $M_e$
with parameters $\bm{\phi}_{e}$ by minimizing
\begin{equation}
\min_{\bm{\phi}_{e} }L_{e}(M_e(\bm{P}_{2D}^{'},\bm{\phi}_{e}), \bm{P}_{3D}^{'}) ,
\label{eq2}
\end{equation}
where $L_{e}$ denotes the loss function between predicted and ground truth 3D poses.

\subsection{Overview of the Proposed Method}
Our proposed framework aims to synthesize new 2D-3D pose pairs from the labelled source domain and the unlabelled target domain, and then use them to train a pose estimator to predict the 3D human pose in the target domain. Among the various factors contributing to the domain gap, camera viewpoints and positions play a significant role by affecting the global positions of the poses. Thus, the key challenge to be met is how to make the augmented data adequately match the distribution of the target dataset. To achieve this, we explore the latent knowledge of the target dataset as follows (also illustrated in Fig. \ref{fig-method}). A pseudo-label generation module $M_p$ is proposed to generate pseudo-3D poses for the target dataset based on the Pose Estimator $M_e$, which is pre-trained on the source domain.
For the {\it source dataset}, we employ an augmentation module, {Pose Augmentor} $M_a$,  that creates more diverse samples to better account for motions and body sizes.
The 2D and 3D poses from both domains are then spatially aligned in a Global Transformation module $M_t$ using a human-centric coordinate system as a bridge.
These aligned 2D and 3D poses are then used to train the Pose Discriminators $M_d^{2D}$, $M_d^{3D}$, and the Pose Estimator $M_e$ jointly, while the discriminators are trained adversarially to maintain both the diversity and plausibility of synthesized data. Finally, the fine-tuned Pose Estimator $M_e$ is applied to $M_p$ to generate updated 3D pseudo-labels of the target dataset again. 
{Since the Global Transformation module is guided by pseudo-labels, updated pseudo-labels enhance alignment between synthesized and target data. Therefore, it may be possible to boost the pose estimation results by iterating the above process. We explore this in our ablations.}

\subsection{Pseudo-Label Generation} \label{sec-pl3dgen}

For the target dataset, we generate 3D pseudo-labels by inferring absolute 3D coordinates of human skeleton joints in the camera coordinate system from the available 2D poses. These 3D pseudo-labels can be obtained using a pre-trained pose estimation method given 2D inputs.

Given a 2D pose $\bm{P}_{2D}^{tgt}=[\bm{x}^{tgt}, \bm{y}^{tgt}]^{T}$ from the target dataset, we obtain the root-relative 3D coordinates $\hat{\bm{P}}_{3D}^{tgt}=[\hat{\bm{X}}^{tgt}, \hat{\bm{Y}}^{tgt},\hat{\bm{Z}}^{tgt}]^{T}$ using the pre-trained pose estimator $M_e$ on the source dataset.
The predicted 3D coordinates of the joints $(\hat{X_{j}}^{tgt}, \hat{Y_{j}}^{tgt},\hat{Z_{j}}^{tgt}), j=\left \{ 0,...,J-1 \right \}$ are root-relative 3D coordinates 
with the root $(\hat{X_{0} }^{tgt}, \hat{Y_{0} }^{tgt},\hat{Z_{0} }^{tgt})$ set to $(0,0,0)$.
To obtain the absolute coordinates of the root, we apply a two-stage optimization process. In the first stage, given the 2D pose and camera intrinsic parameters $\bm{K}$, the 3D pose can be inferred by assuming an initial depth value.

The corresponding bone lengths are then derived from the estimated 3D pose. To ensure anatomical consistency, we minimize the discrepancy between the estimated bone lengths and their averaged values using least squares, 
and thus, we obtain the 3D absolute coordinates of all joints $ \tilde{\bm{P}}_{j}=({\tilde{X}_{j} }^{tgt}, {\tilde{Y}_{j} }^{tgt},{\tilde{Z}_{j} }^{tgt}), j=\left \{ 0,...,J-1 \right \}$. {In the second stage, we minimize the error between reprojected 2D pose from $\hat{\bm{P}} _{3D}^{tgt}+\bm{P}_{r}$ and real 2D pose $\bm{P}_{2D}^{tgt}$ using the least squares method again, where $\bm{P}_{r}$ is the absolute coordinates of the root joint and its initial value is set to root joint $\tilde{\bm{P}}_{0}$.}
Finally, we arrive at the estimated absolute 3D coordinates of all joints $\tilde{\bm{P}}_{3D}^{tgt}$ as the 3D pseudo-labels of the target dataset.

There are no training parameters in $M_p$, while the absolute 3D coordinates are obtained by applying bone-length and 2D projection constraints through the least squares optimization process. Based on the above pseudo-label generation method, we generate 3D pseudo-labels that approximate the target distribution, exploiting the latent knowledge of the target dataset. These pseudo-labels play a pivotal role in guiding the subsequent global transformation process.

\begin{figure}[t]
    \centerline{\includegraphics[width=8.4cm]{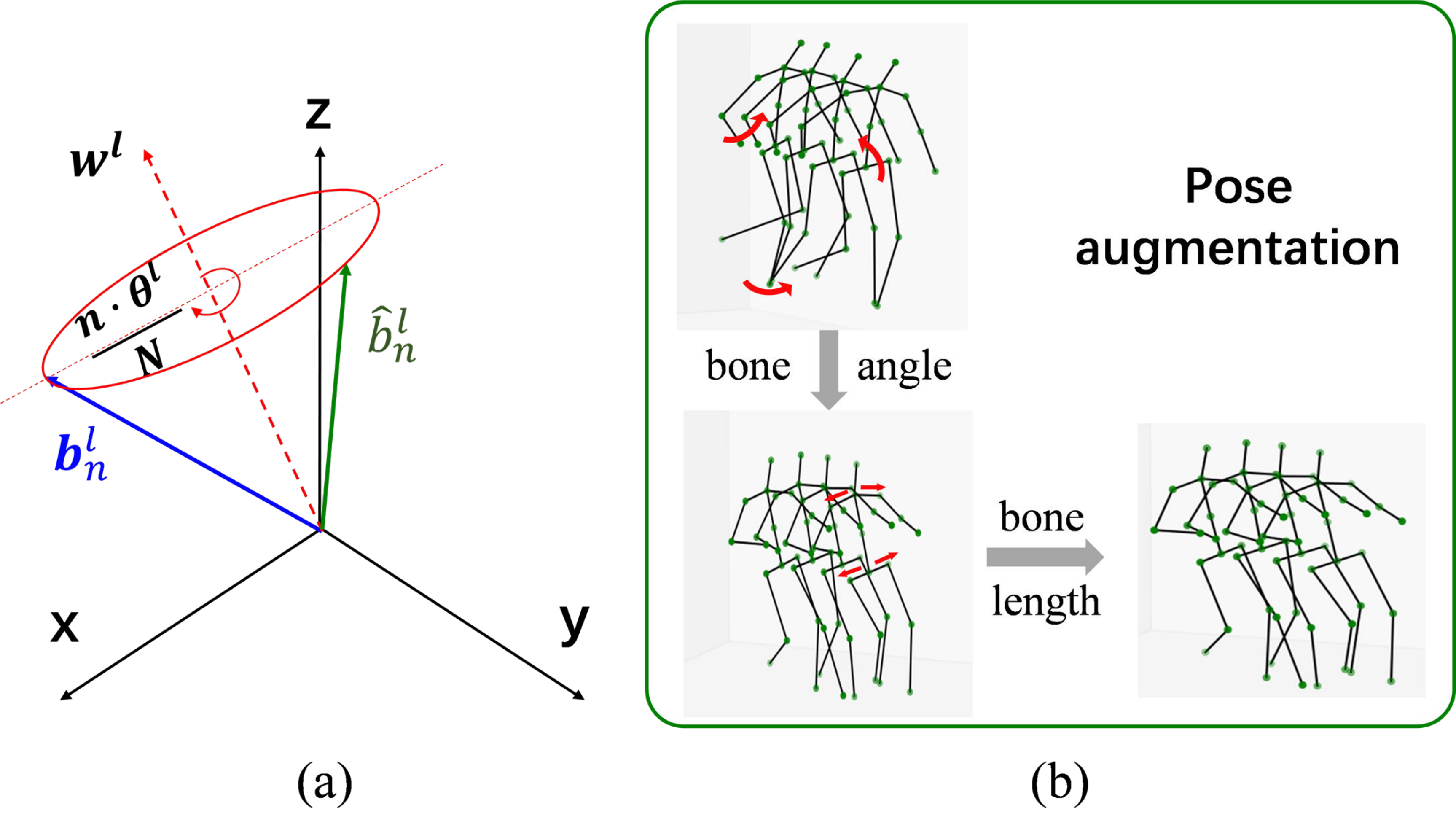}}
    \caption{\bf (a) Bone angle augmentation for a 3D pose sequence. (b) A source 3D pose sequence is augmented by modifying its motion via bone angle operation and body size via bone length operation.}
    \label{fig-BL}
\end{figure}

\subsection{Pose Augmentor}
\label{sec-PAadaptpose}

The poses in the source dataset are augmented using the bone augmentation process from  AdaptPose \cite{gholami2022adaptpose}. This enables our Pose Augmentor $M_a$ to modify bone angles across frames, introducing greater motion diversity. As a result, the augmented poses capture a wider range of actions.
Additionally, bone lengths are adjusted to accommodate variations in human body proportions, enhancing the model's adaptability to different body sizes.

Given a sequence of 3D poses $\left \{ \bm{P}_{3D}^{src}  \right \} _{n=0}^{N-1}$ from the source dataset, our augmentor converts them into their {corresponding bone vectors $\left \{ \bm{B}^{src}  \right \} _{n=0}^{N-1}\in R^{N\times (J-1)\times 3}$. Each original bone vector $b^l_n \in R^{1\times 3}$ (i.e., the $l$-th bone vector from $\bm{B}^{src}$ in $n$-th frame) is rotated about an axis $\bm{w}^l \in R^{1\times 3}$ by a certain rotation angle $\theta^l$ proportional to the number of frames.} This guarantees that the bone changes are temporally plausible and consistent. Additionally, to adjust the bone length, the pose augmentor scales it using a ratio parameter $\lambda^l$, which systematically modifies the bone length while preserving the structural coherence of the motion.

{To predict the parameters ($\bm{w}\in R^{(J-1)\times 3},\theta\in R^{(J-1)\times 1},\lambda\in R^{(J-1)\times 1}$), following \cite{gholami2022adaptpose}, the input is formed by concatenating a Gaussian-noise vector with the 3D pose of the middle frame as the representative of the whole sequence. The Gaussian noise introduces randomness and diversity for feature extraction. Subsequently, {four} fully connected layers {followed by batch normalization and a leaky ReLU}
are used to extract features and learn  these  parameters  \cite{gholami2022adaptpose}.}

{An example of the above process is shown in Fig. \ref{fig-BL}(a), where by rotating $ \bm{b}_{n}^{l}$  along the axis $\bm{w}^l$ for $\frac{n\cdot \theta^l}{N}$ degrees, we obtain the augmented bone vectors $\hat{\bm{b}}_{n}^{l}$ with revised joint angles. Then, we apply the bone length ratio $\lambda ^ l$ to obtain $(1+\lambda ^l)\cdot \hat{\bm{b}}_{n}^{l}$ as the new bone vector.
All the new bone vectors $\hat{B}\in R^{N\times (j-1)\times 3}$, where $(1+\lambda ^l)\cdot \hat{\bm{b}}_{n}^{l}$ is the $l$-th bone vector of $n$-th frame, are then converted back to generate the augmented 3D pose sequence $\left \{  \bm{P}_{3D}^{\mathcal{A}}\right \} _{n=0}^{N-1}$. Fig. \ref{fig-BL}(b) illustrates examples of the effects of the bone angle and bone length augmentation.}

\begin{figure}[t]
   \includegraphics[width=7.8cm]{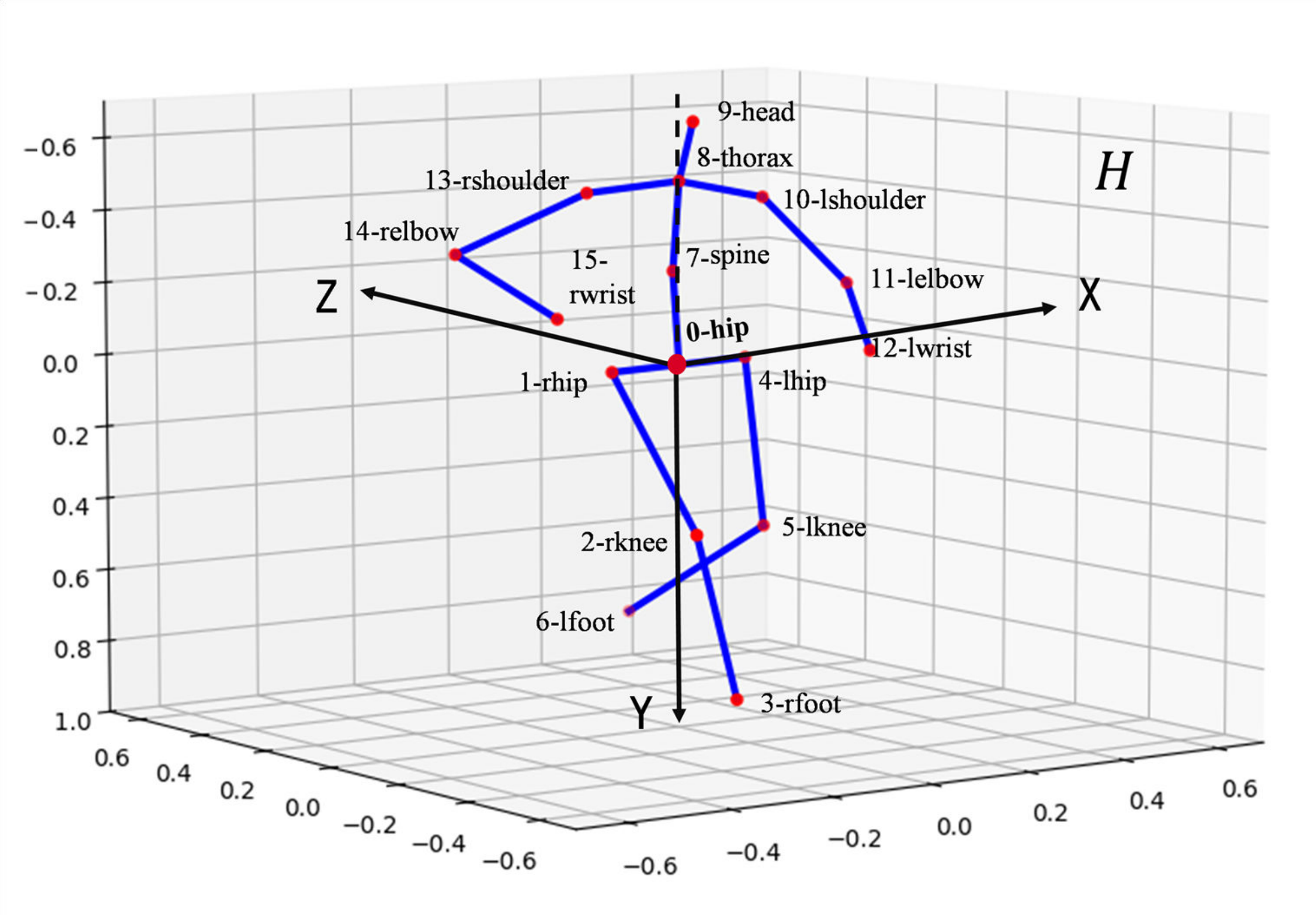}
   \caption{{\bf Human skeleton joints and human-centric coordinate system.} Where relevant, each joint's label is preceded with `r' and `l' for right and left sides. }
   \label{fig-coordinate}
\end{figure}

\subsection{Global Transformation}
\label{sec-gt}

Camera viewpoints and positions across datasets and/or domains lead to significant discrepancies in the 3D absolute coordinates of human poses. Given a 3D human pose $P^{A}$ captured from a source domain and another 3D pose $P^{B}$ from a target domain, the objective would be to eliminate the domain discrepancy by transforming $P^{A}$ so that it aligns with $P^{B}$. This problem can be formulated as an optimal rigid alignment task, where the goal is to estimate the rotation matrix $R$ and translation matrix $T$ to minimize $ \left\| R\cdot P^{A}+T-P^{B}\right\|$.

A well-established solution to this is the Kabsch algorithm \cite{kabsch1976solution, umeyama2002least}, which computes the cross-covariance matrix between the two point sets,  followed by Singular Value Decomposition to derive the optimal transformation. However, this process is complex and time-consuming, especially when applied to the training process on large datasets.

Inspired by prior works \cite{wang2020predicting, wei2019view} which introduced a body-centered coordinate system to achieve viewpoint invariance, we propose to use a human-centric coordinate system $H$ as a bridge to align 3D coordinates across domains.
A 3D human pose $P$ can be expressed as $P=P_r+\Delta P$, where $P_r$ denotes the 3D absolute coordinate of the root which usually is the middle hip joint, and $\Delta P$ is the relative coordinates compared to the root. The transformation of $P^{A}$ to align with $P^{B}$ incorporates three steps.
The first step is to center $P^{A}$ at $H$ by subtracting the root's location $P_r^A$. In the second step, the aim is to find a rotation to minimize the error between joints' relative coordinates. In the third step, a translation is applied to align the root joint with $P_r^B$. The theoretical derivations can be found in Appendix A.

As can be seen  in Fig. \ref{fig-coordinate}, the origin of the human-centric coordinate system is taken as the center of the left and right hip joints (labeled 0-hip). The x-axis is defined as the unit vector pointing from the right hip to the left hip, while the x-y plane is defined by three joints: the right hip, left hip, and the thorax joint. The z-axis is determined by the unit vector perpendicular to x-y plane and y-axis is the cross-product of x-axis and z-axis.

In the absence of ground-truth 3D poses in the target domain, we adopt the pseudo-labels generated in Section \ref{sec-pl3dgen} as surrogate {ground-truth} labels that will be used to guide the global transformation process.
Given an augmented 3D pose $\bm{P}_{3D}^{\mathcal{A}}$  from the source dataset, we aim to transform it using rotation and translation matrices to align with a 3D pseudo-pose $\tilde{\bm{P}}_{3D}^{tgt}$ sampled from the target dataset. 
First, $\bm{P}_{3D}^{\mathcal{A}}$ is centered at $H$, subtracting the root  location $\bm{P}_{r}^{\mathcal{A}}$.
Second, a rotation is applied to align the joints' relative coordinates. Using the human coordinate system $H$ as a bridge, the rotation $\bm{R}$ from the source pose to the target pose can be decomposed to be (i) from the source camera coordinate system $C_s$ to $H$ and then (ii) from $H$ to the target camera coordinate system $C_t$. Third, a translation is applied to align the root of target pose $\tilde{\bm{P}}_{r}^{tgt}$. The entire transformation is 
\begin{equation}
\bm{P}_{3D}^{'}=\bm{R}_{2}^{-1}\cdot \bm{R}_{1}\cdot (\bm{P}_{3D}^{\mathcal{A}}- \bm{P}_{r}^{\mathcal{A}} ) + \tilde{\bm{P}}_{r}^{tgt},
\label{eq_RT3}
\end{equation}
\noindent where $\bm{R}_{1}$ and $\bm{R}_{2}$ denote the rotation matrices from $C_s$ to $H$ and $C_t$ to $H$ respectively, and ${\bm{P}}_{3D}^{'}=[\bm{X}^{'}, \bm{Y}^{'}, \bm{Z}^{'}] ^{T} \in R^{3\times J}$ is the transformed 3D pose. See Appendix B for the details of computing the rotation matrix from a camera coordinate system to $H$.

\begin{figure}[!t]
    \centerline{\includegraphics[width=8.2cm]{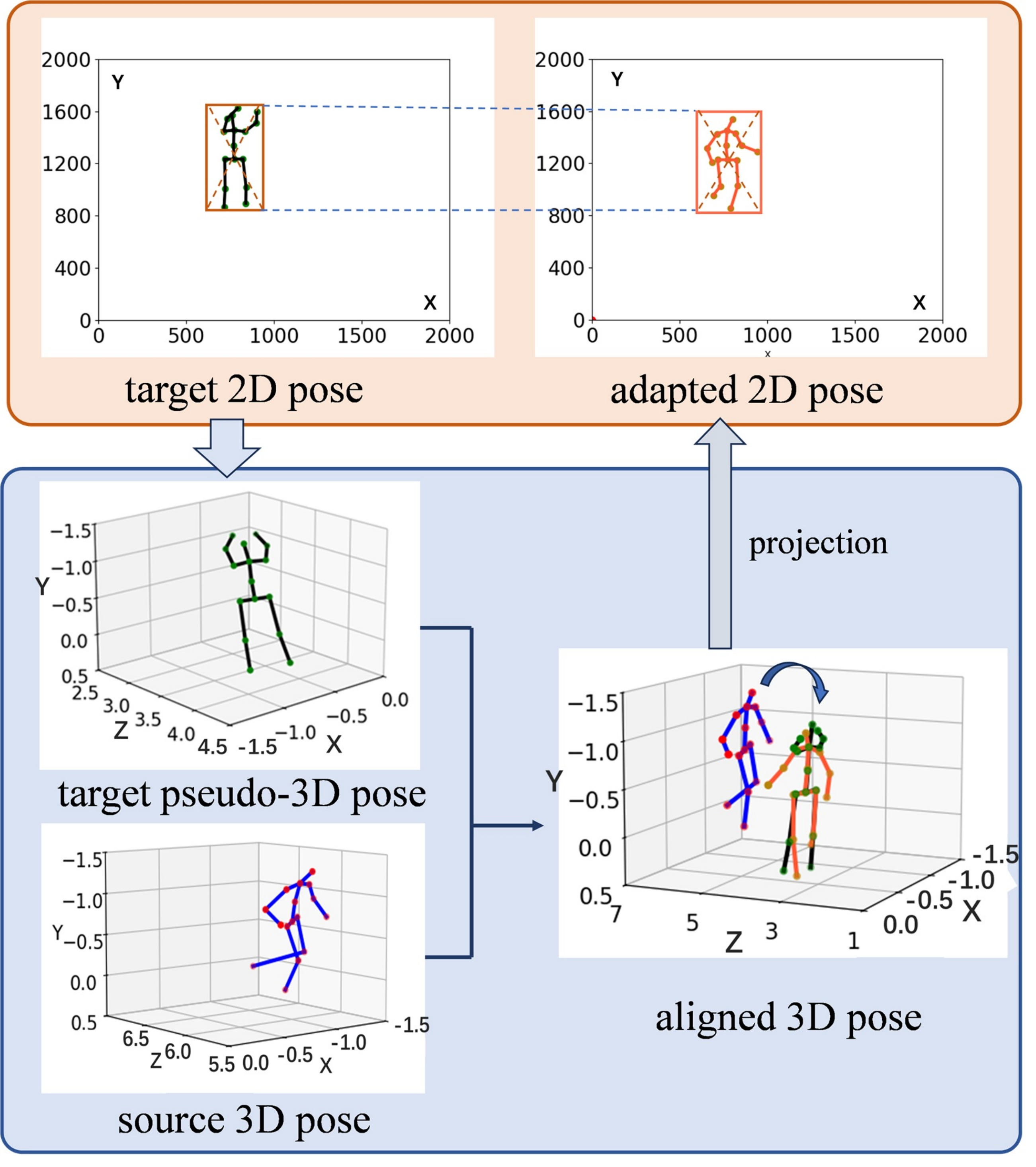}}
    \caption{{\bf Visualization of the global transformation.} A pseudo-3D pose is inferred from the target 2D pose. It is used to guide the rotation and translation of the source 3D pose for alignment. Given the aligned 3D poses, corresponding 2D poses are obtained through projection using the intrinsic camera parameters of the target dataset so that the 2D root positions are also aligned.}
    \label{fig-visualization}
\end{figure}

Finally, we can project ${\bm{P}}_{3D}^{'}$ into 2D pose ${\bm{P}}_{2D}^{'}=[\bm{u},\bm{v}]^{T} \in R^{2\times J}$ where $(\bm{u},\bm{v})$ denote the 2D image coordinates, obtained as
\begin{equation}
\begin{array}{l}
u_j=f_{x}\cdot \frac{X_j^{'} }{Z_j^{'} }  +c_{x}, \\[10pt]
v_j=f_{y}\cdot \frac{Y_j^{'} }{Z_j^{'} }  +c_{y},  ~ j=\left \{ 0,...,J-1 \right \} ,
\end{array}
\label{eq4}
\end{equation}
where 
$(f_{x}, f_{y}, c_{x}, c_{y})$ are the target dataset camera's intrinsic parameters.
As shown in Fig. \ref{fig-visualization}, after global transformation, 3D poses from the source and target domains are aligned. 
For 2D poses, their root positions are also aligned and the scales are similar. This {alignment} process 
significantly reduces the domain discrepancies caused by differences in camera viewpoints and positions. We analyse the effect of the proposed human-centered global transformation module compared with the Kabsch algorithm \cite{kabsch1976solution,umeyama2002least} in Sec. \ref{Diff_GT}.

\subsection{Iterative training architecture}
\label{sec-iter}
After obtaining the aligned 2D and 3D poses ${\bm{P}}_{2D}^{'}$ and ${\bm{P}}_{3D}^{'}$ respectively, we train an adversarial pose augmentation framework to ensure the fidelity of the generated human poses (following previous works \cite{gong2021poseaug,chai2023global}).
This augmentation framework involves our generator, i.e., the pose augmentor and the discriminators $M_d^{2D}$, $M_d^{3D}$.
{The generator tries to generate fake samples as close as possible to real samples, while the discriminator tries to distinguish between them.} $M_d^{2D}$ uses the aligned 2D pose ${\bm{P}}_{2D}^{'}$ and the target 2D pose ${\bm{P}}_{2D}^{tgt}$ at the middle frame as input, to adversarially guide ${\bm{P}}_{2D}^{'}$ to align with the target sample. This is unlike PoseAug \cite{gong2021poseaug} which uses the 2D poses of source dataset as input. The discriminator $M_d^{3D}$ uses the middle-frame 3D augmented pose ${\bm{P}}_{3D}^{\mathcal{A}}$ and the source pose ${\bm{P}}_{3D}^{src}$ as input to ensure the fidelity of poses after bone angle and bone length augmentation. The middle frame is taken as the input since nearby frames are expected to show minor variations, making the discriminator effective across the temporal window.

Inspired by Kinematic Chain Space (KCS) \cite{wandt2018kinematic,wandt2019repnet}, we use the KCS representation of 3D pose as input rather than 3D joints.
KCS is a representation of human pose that captures the structural relationships between bones by computing the inner products of bone vectors. Instead of considering the entire body pose as a whole, we use a part-aware KCS \cite{gong2021poseaug}  which separates the whole body into 5 parts: torso and left/right arm/leg. By focusing on local joint angles, this enlarges the feasible region of the augmented pose while maintaining plausibility and promoting diversity. To compute the part-aware KCS of an input pose $X$, we convert the pose to its bone direction vector $b$ first. Then, $b$ is separated corresponding to the five parts, denoted as $b_i, i=1,...,5$. The local joint angle matrix  $\mathcal{K}_i$  is calculated for each part as
\begin{equation}
\mathcal{K}_i=b_i^T \cdot b_i ,
\label{eq-kcs}
\end{equation}
which captures the inter-joint angle relationships within the $i$-th part. $\mathcal{K}_i$ is taken as the input to the 3D pose discriminator $M_d^{3D}$.

In each epoch, the generator $M_a$, discriminators $M_d^{2D}$, $M_d^{3D}$ and the pose estimator $M_e$ are trained jointly. $M_e$ is the pose estimation model that is fine-tuned to adapt to the target dataset.
Their loss functions are as follows.

\noindent \textbf{Discriminator loss.} We use least squares GAN (LSGAN) \cite{mao2017least} to train the adversarial pose augmentation framework:
\begin{equation}
\begin{array}{l}
L_{dis}=  L_{dis}^{2D} + L_{dis}^{3D}, \\
L_{dis}^{2D}=\frac{1}{2} \mathbb{E}[M^{2D}_{d}(\bm{P}_{2D}^{tgt})-1)^{2}]+ \frac{1}{2} \mathbb{E}[M^{2D}_{d}(\bm{P}_{2D}^{'})^{2}],  \\
L_{dis}^{3D}=\frac{1}{2} \mathbb{E}[(M^{3D}_{d}(\bm{P}_{3D}^{src})-1)^{2}]+ \frac{1}{2} \mathbb{E}[M^{3D}_{d}(\bm{P}_{3D}^{\mathcal{A}})^{2}].
\end{array}
\label{eq5}
\end{equation}
\textbf{Generator loss.} {The adversarial loss of the generator $M_a$ is the feedback from the discriminators:}
\begin{equation}
\begin{array}{l}
L_{{adv} }=\alpha \cdot L_{adv}^{2D}+\beta \cdot L_{adv}^{3D} , \\
L_{adv}^{2D}=\frac{1}{2} \mathbb{E}[(M^{2D}_{d}(\bm{P}_{2D}^{'})-1)^{2}]+ \frac{1}{2} \mathbb{E}[M^{2D}_{d}(\bm{P}_{2D}^{tgt})^{2}],  \\
L_{adv}^{3D}=\frac{1}{2} \mathbb{E}[(M^{3D}_{d}(\bm{P}_{3D}^{\mathcal{A}})-1)^{2}]+ \frac{1}{2} \mathbb{E}[M^{3D}_{d}(\bm{P}_{3D}^{src})^{2}].
\end{array}
\label{eq6}
\end{equation}
\textcolor{black}{As shown in Eqs. \ref{eq5} and \ref{eq6}, the generator is trained to minimize the discrepancy between the aligned 2D pose $\bm{P}_{2D}^{'}$ and the target 2D pose $\bm{P}_{2D}^{tgt}$, as well as between the augmented 3D pose $\bm{P}_{3D}^{\mathcal{A}}$ and the source 3D pose $\bm{P}_{3D}^{src}$. In contrast, the discriminator aims to distinguish between these respective pose pairs. In this adversarial approach, the 3D pose discriminator $M^{3D}_{d}$ is used for evaluating the joint angle plausibility since the KCS representation encapsulates the inter-joint angle information. As the 2D poses contain
information such as view point (rotation), position (translation), and body size (bone length), the 2D pose discriminator $M^{2D}_{d}$ is used for evaluating the body size, camera viewpoint, and position plausibility \cite{gong2021poseaug}. }

\noindent\textbf{Pose estimation loss.} The standard Mean Squared Error
(MSE) loss is employed for the Pose Estimator $M_e$,
\begin{equation}
L_{S}=\left \| \bm{P}-\tilde{\bm{P}} \right \| _{2}^{2},
\label{eq7}
\end{equation}
where $\bm{P}$ and $\tilde{\bm{P}}$ are the ground truth and predicted 3D poses respectively.
Before training $M_e$, the generator and the discriminators need to be warmed up for $m$ epoch.

After the training process, we obtain a fine-tuned Pose Estimator $M_{e}$ which generates predicted 3D poses $\hat{\bm{P}} _{3D}^{tgt}$ for the target dataset. These poses could be used as input for the Pseudo-Label Generation module $M_p$ again, and thus, {the training process can be iterative and the number of iterations is studied in the ablation study.}

\section{Experiments}

\subsection{Datasets and Metrics}
Following previous works, we evaluate our method on several
widely used large-scale 3D human pose estimation benchmarks, including MPI-INF-3DHP \cite{mehta2017vnect}, 3DPW \cite{von2018recovering}
and Human3.6M \cite{ionescu2013human3}.

\textbf{MPI-INF-3DHP (3DHP)} is a large-scale, in-the-wild 3D human pose dataset. It includes 2D and 3D data from eight subjects performing eight different activities, recorded in both indoor and outdoor scenes. 3DHP is ideal for evaluating methods that aim to generalize to real-world scenarios. We follow previous works \cite{kolotouros2019learning,gong2021poseaug} by using the test set, which consists of approximately 2,929 frames. Evaluation is conducted using three main metrics: Mean Per Joint Position Error (MPJPE), Percentage of Correct Keypoints (PCK) with a 150mm threshold, and Area Under the Curve (AUC) calculated over various PCK thresholds.

\textbf{3D Poses in the Wild Dataset (3DPW)} offers a unique and challenging benchmark for 3D human pose estimation in a variety of uncontrolled outdoor settings. It includes dynamic camera motion and natural lighting conditions, with scenes recorded at 25 fps.
In evaluations, MPJPE and Procrustes-Aligned Mean Per Joint Position Error (PA-MPJPE) are used as evaluation metrics.

\textbf{Human3.6M (H3.6M)} is an extensive dataset for 3D human pose estimation, captured in a controlled indoor environment. The dataset includes over 3.6 million frames, covering seven main subjects (S1, S5, S6, S7, S8, S9, and S11) engaged in 15 diverse activities.
Following previous works \cite{chai2023global,peng2024dual}, H3.6M has two main settings:
(i) for cross-dataset evaluation, we use subjects S1, S5, S6, S7, and S8 as the source domain, with the other dataset as the target domain, and (ii) for cross-scenario evaluation, we use only S1 as the source domain, with S5, S6, S7, and S8 as the target domain. Measures MPJPE and PA-MPJPE are standard metrics employed to evaluate performance across these tasks.

\subsection{ Implementation Details}
\textbf{Pose Estimator.}
Following previous works \cite{gong2021poseaug,gholami2022adaptpose,liu2023posynda}, we use single-frame-based VideoPose3D \cite{pavllo20193d} and video-based MixSTE \cite{zhang2022mixste} for our Pose Estimator $M_e$. The input sequence length of MixSTE was set to 27 for 3DHP and 3DPW, and 243 for Human3.6M following the settings in \cite{liu2023posynda}.
The pre-trained weights from the source dataset are used as initial weights. We adopt the 16-keypoint human model with the 0-th joint (hip center point) as the origin and we use ground truth 2D keypoints as input.

\textbf{Generators and discriminators.}
The generator $M_a$ has three 3-layer residual MLPs with LeakyReLU to generate augmented bone angles and bone lengths. We set $\alpha=1$ and $\beta=6$ 
in Eq. \ref{eq6}. The discriminators $M_d^{2D}$ and $M_d^{3D}$ use similar architectures for processing 2D and 3D poses. $M_d^{2D}$ incorporates 4-layer residual MLPs with LeakyReLU, taking the 2D keypoints as input.  $M_d^{3D}$ consists of five 4-layer residual MLPs with LeakyReLU with each corresponding to a part of KCS representation.

\textbf{Training.}
The initial learning rates for the discriminators, generator, and pose estimator are all set to 0.0001. They all use Adam optimizer.
The experiments are conducted with a batch size of 1024 for 50 epochs and there is a warm-up phase lasting 2 epochs to train the generator and discriminators.

\subsection{Results}

\textbf{Cross-dataset Evaluation.}
For cross-dataset evaluations, the source and target data exhibit large domain gaps in terms of camera settings, human body size, and motions. We perform three cross-dataset experiments.

In the first experiment, H3.6M is the source and 3DHP is the target dataset (see Table \ref{tab1}).  For single frame-based VideoPose3D \cite{pavllo20193d} as the Pose Estimator $M_e$, {our method outperforms PoSynDA\cite{liu2023posynda} by 2.5mm$\downarrow$ on MPJPE and PoseDA\cite{chai2023global} by 2.1$\uparrow$ on AUC.
For video-based MixSTE \cite{zhang2022mixste} as the Pose Estimator, our method improves on PoSynDA\cite{liu2023posynda} by 4.2mm$\downarrow$ in MPJPE, 0.9\%$\uparrow$ in PCK and 5.4$\uparrow$ in AUC.}
{Note, our method using MixSTE \cite{zhang2022mixste} surpasses the target-specific, fully supervised MixSTE \cite{zhang2022mixste} method itself (from 54.9mm to 54.0mm in MPJPE).}

\begin{table}[t]
\setlength{\tabcolsep}{3.9pt}
\centering
\caption{{\bf Cross-dataset evaluation results on target 3DHP with source H3.6M.} First two methods marked with * were fully-supervised, i.e., trained and tested on 3DHP. `Vid' denotes method is video-based with $T=27$.  VideoPose3D\cite{pavllo20193d} (VP3D) and MixSTE \cite{zhang2022mixste} are indicated as the Pose Estimator used in each method.
}
\footnotesize
\scalebox{0.85}{
\begin{tabular}{ll|c|ccc}
\toprule
{\bf Method}    &            & {\bf Vid} & {\bf MPJPE $\downarrow$} & {\bf PCK$\uparrow$}  & {\bf AUC$\uparrow$ }  \\ \hline
SimpleBaseline \cite{martinez2017simple} * & ICCV2017 &   & 84.3  & 85.0 & 52.0 \\
MixSTE \cite{zhang2022mixste} *  & CVPR2022  &  & 54.9  & 94.4 & 66.5 \\ \hline
PoseAug (VP3D)\cite{gong2021poseaug}   &CVPR2021 &  & 73.0  & 88.6 & 57.3 \\
PoseDA (VP3D)  \cite{chai2023global}  &ICCV2023&  & 61.3  & 92.1 & \underline{62.5}\\
PoSynDA (VP3D)  \cite{liu2023posynda}  &ACMMM2023  &   & \underline{60.2}  & \textbf{93.1} & 58.4 \\
Dual-Augmentor (VP3D) \cite{peng2024dual}  &CVPR2024 &   & 63.1  & \underline{92.9} & 60.7 \\
\textbf{Ours (VP3D)}   &   &  & \textbf{57.7}  & 92.5 & \textbf{64.6} \\ \hline
AdaptPose (VP3D) \cite{gholami2022adaptpose}  &CVPR2022 &  $\checkmark$ & 68.3  & 90.2 & 59.0 \\
PoSynDA (MixSTE)  \cite{liu2023posynda}  &ACMMM2023  &$\checkmark$ & \underline{58.2} &\underline{93.5} & \underline{59.6} \\
\textbf{Ours (MixSTE)}   & &$\checkmark$ & \textbf{54.0}  & \textbf{94.4} & \textbf{65.0} \\
\bottomrule
\end{tabular}}
\label{tab1}
\end{table}

In the second experiment, H3.6M is the source and 3DPW is the target dataset (see Table \ref{tab2}). For both single-frame and video-based methods, our approach achieves state-of-the-art results. Amongst single-frame methods, ours surpasses  PoseDA\cite{chai2023global} in MPJPE by 12.0mm$\downarrow$ and in PA-MPJPE by 10.2mm$\downarrow$ respectively. For video-based approaches, again in comparison with the state-of-the-art PoSynDA\cite{liu2023posynda}, we improve by 8.9mm$\downarrow$ and 3.5mm$\downarrow$ in MPJPE and PA-MPJPE, respectively.
Our method manages greater strides in improved results on 3DPW than on 3DHP, as it is designed to address the issue of variations in camera position and orientation which are more pertinent to the dynamic camera motion scenes in 3DPW.

\begin{table}[t]
\setlength{\tabcolsep}{3.9pt}
\centering
\caption{{\bf Cross-dataset evaluation results on target 3DPW with source H3.6M.} First two methods marked with * were fully-supervised, i.e., trained and tested on 3DPW. `Vid' denotes method is video-based with $T=27$.  VideoPose3D\cite{pavllo20193d} (VP3D) and MixSTE \cite{zhang2022mixste} are indicated as the Pose Estimator used in each method.}
\scalebox{0.85}{
\begin{tabular}{ll|c|cc}
\toprule
\bf{Method}    &   & \bf{Vid} & \bf{MPJPE$\downarrow$} & \bf{PA-MPJPE$\downarrow$ } \\ \hline
VIBE \cite{kocabas2020vibe} * &CVPR2020     & $\checkmark$  & 82.9                           & 51.9                      \\
Lin et al. \cite{lin2021mesh} * &ICCV2021 &    & 74.7                        & 45.6                   \\  \hline
PoseAug (VP3D) \cite{gong2021poseaug} &CVPR2021 &   &94.1   & 58.5                       \\
PoseDA (VP3D) \cite{chai2023global}&ICCV2023   &   & \underline{ 87.7}                          & \underline{55.3}    \\
Dual-Augmentor (VP3D)  \cite{peng2024dual} &CVPR2024   &   &106.6            & 73.2                    \\
\textbf{{Ours (VP3D)}}   &   &   & \textbf{75.7}& \textbf{45.1}  \\ \hline
VIBE \cite{kocabas2020vibe}  &CVPR2020   & $\checkmark$  & 93.5                         &56.5                    \\
BOA  \cite{guan2021bilevel} &CVPR2021   & $\checkmark$  &77.2                        &  49.5                  \\
AdaptPose (VP3D)  \cite{gholami2022adaptpose}&CVPR2022 & $\checkmark$  & 81.2            & 46.5                    \\

PoSynDA (MixSTE) \cite{liu2023posynda}  &ACMMM2023 & $\checkmark$  & \underline{75.5}  & \underline{45.4}                      \\
\textbf{{Ours (MixSTE)}}   &   & $\checkmark$  & \textbf{66.6}& \textbf{41.9}  \\
\bottomrule
\end{tabular}}
\label{tab2}
\end{table}

In the third experiment, we demonstrate the effectiveness of our method given a challenging viewpoint. We add to the 3DHP dataset a simulated virtual camera $C_v$, positioned at 2m above the ground plane at a depression angle of $45^{\circ}$. For ease of exposition, we call this dataset, 3DHP-C. We compute new 3D poses under the coordinate system of the virtual camera
 and then use the intrinsic parameters of the original camera in 3DHP dataset to compute 2D poses,  such that
\begin{equation}
\begin{array}{l}
\begin{bmatrix} X_{new} \\Y_{new} \\Z_{new} \end{bmatrix}=\begin{bmatrix}
 1 & 0 & 0 \\
 0 & \cos(45^{\circ } )  & \sin(45^{\circ } )  \\
 0 & \sin(45^{\circ })  & \cos(45^{\circ } )
\end{bmatrix}\cdot \begin{bmatrix} X \\Y+2 \\Z \end{bmatrix},\\
\begin{bmatrix}u_{new}
\\v_{new}
 \\1
\end{bmatrix}=\begin{bmatrix}
 f_{x}^{p}& 0 & c_{x}^{p} \\
 0& f_{y}^{p} & c_{y}^{p} \\
0 &0  &1  \\
\end{bmatrix}\cdot \begin{bmatrix}
\frac{X_{new}}{Z_{new}} \\[3pt]
 \frac{Y_{new}}{Z_{new}}\\[3pt]
1 \end{bmatrix} ,
\end{array}
\label{eq8}
\end{equation}
{where $(X, Y, Z)$ is the 3D coordinate of a joint in 3DHP, $(X_{new}, Y_{new}, Z_{new})$ is the 3D coordinate of the joint in 3DHP-C after processing,  ($(f_{x}^{p},f_{y}^{p},c_{x}^{p},c_{y}^{p} )$ are the intrinsic parameters of the original camera in 3DHP, and $(u_{new}, v_{new})$ is the 2D coordinate of the joint in 3DHP-C.}

Table \ref{tab3} presents comparative results against PoseAug\cite{gong2021poseaug} and AdaptPose\cite{gholami2022adaptpose} with H3.6M as the source and 3DHP-C as the target dataset.
{We do not compare with PoseDA \cite{chai2023global} as their complete codes are not available, nor with PoSynDA \cite{liu2023posynda} since we obtained excessively incorrect results using their provided code.}
Our method outperforms by a large margin. This improvement stems from our explicit handling of camera viewpoint discrepancies which impresses the potential effectiveness of the proposed method {under challenging viewpoints.}

\textbf{Cross-scenario Evaluation.}
In cross-scenario evaluations,  subject S1's data in H3.6M is the source and subjects S5, S6, S7, and S8's data in H3.6M are the target.
As shown in Table \ref{tab4}, among related work, our method performs slightly better in MPJPE, and (almost) as well in PA-MPJPE. In this experiment, the camera parameters do not change much between subjects.

\begin{table}[!h]
\centering
\caption{{\bf{Cross-dataset evaluation results on target 3DHP-C with source H3.6M.} }VideoPose3D\cite{pavllo20193d} (VP3D) and MixSTE \cite{zhang2022mixste} are indicated as the Pose Estimator used in each method.}
\scalebox{0.9}{
\begin{tabular}{ll|ccc}
\toprule
\bf{Method}    &  & \bf{MPJPE$\downarrow$} & \bf{PCK$\uparrow$ }  & \bf{AUC$\uparrow$}   \\ \hline
PoseAug (VP3D)\cite{gong2021poseaug} &CVPR2021 & 145.4 & 61.2 & 33.3 \\
AdaptPose (VP3D) \cite{gholami2022adaptpose} &CVPR2022 &{132.8} &{65.0} &{35.6} \\
\textbf{{Ours (VP3D)}}   &     & \underline{77.3}  & \underline{87.6} & \underline{54.5} \\
\textbf{{Ours (MixSTE)}}   &     & \textbf{57.8}  & \textbf{92.0} & \textbf{64.1} \\
\bottomrule
\end{tabular}}
\label{tab3}
\end{table}

\begin{table}[!htp]
\centering
\caption{{\bf{Cross-scenario evaluation results on target H3.6M (S5,S6,S7,S8) with source H3.6M (S1).}} `Vid' denotes method is video-based  with $T=243$. VideoPose3D\cite{pavllo20193d} (VP3D) and MixSTE \cite{zhang2022mixste} are indicated as the Pose Estimator  used in each method.}
\scalebox{0.8}{
\begin{tabular}{ll|c|cc}
\toprule
\bf{Method}  &    & \bf{Vid} & \bf{MPJPE$\downarrow$} & \bf{PA-MPJPE$\downarrow$} \\ \hline
VideoPose3D \cite{pavllo20193d} &CVPR2019 &   & 64.7  & --       \\
Li et al. \cite{li2020cascaded} &CVPR2020 &   & 62.9  & --       \\
PoseAug (VP3D) \cite{gong2021poseaug}   &CVPR2021 &   & 56.7  & --       \\
PoseDA (VP3D) \cite{chai2023global} &ICCV2023    &   & \underline{49.9}  & \textbf{34.2}     \\
Dual-Augmentor (VP3D) \cite{peng2024dual}&CVPR2024 &   & 50.3  & --     \\
\textbf{{Ours (VP3D)}}    &    &   & \textbf{49.7}  & \textbf{34.2}     \\ \hline
AdaptPose (VP3D) \cite{gholami2022adaptpose} &CVPR2022 & $\checkmark$ & 54.2  & 35.6     \\
PoSynDA (MixSTE) \cite{liu2023posynda} &ACMMM2023   & $\checkmark$  & \underline{48.1}  & \textbf{33.2}     \\
\textbf{{Ours (MixSTE)}}    &    &  $\checkmark$ & \textbf{47.9}  & \underline{33.5}    \\
\bottomrule
\end{tabular}}
\label{tab4}
\end{table}

\begin{figure*}[!htp]
  \centering
   \includegraphics[width=16cm]{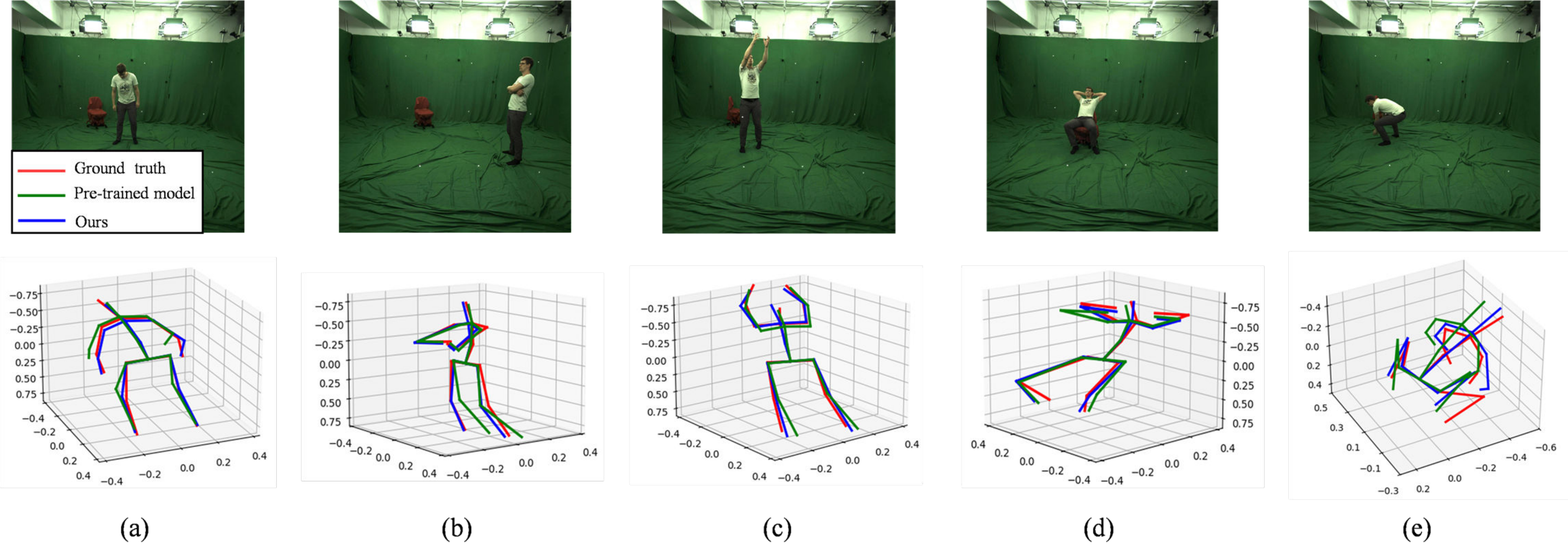}
   \caption{{\bf{Example 3D pose estimations on 3DHP dataset.}} (a)-(d) Our method (blue) performs significantly better than the pre-trained model (green) with smaller errors compared with the ground truth (red). (e) {In cases of severe self-occlusion, our method fails to obtain sufficient estimations. }
   }.

   \label{fig-compare}
\end{figure*}

\textbf{Data Visualizations.} Fig. \ref{fig-compare} (a)-(d) {presents} example qualitative results on the 3DHP dataset which show that compared to the pre-trained baseline model VideoPose3D\cite{pavllo20193d}, our method performs significantly better, especially for joints that are farther away from the root joint. Fig. \ref{fig-compare}(e) illustrates a failure case of our method where the estimation of joints on the right side of the body is not accurate enough due to extreme self-occlusion.

\begin{figure}[!htp]
  \centering
   \includegraphics[width=7.8cm]{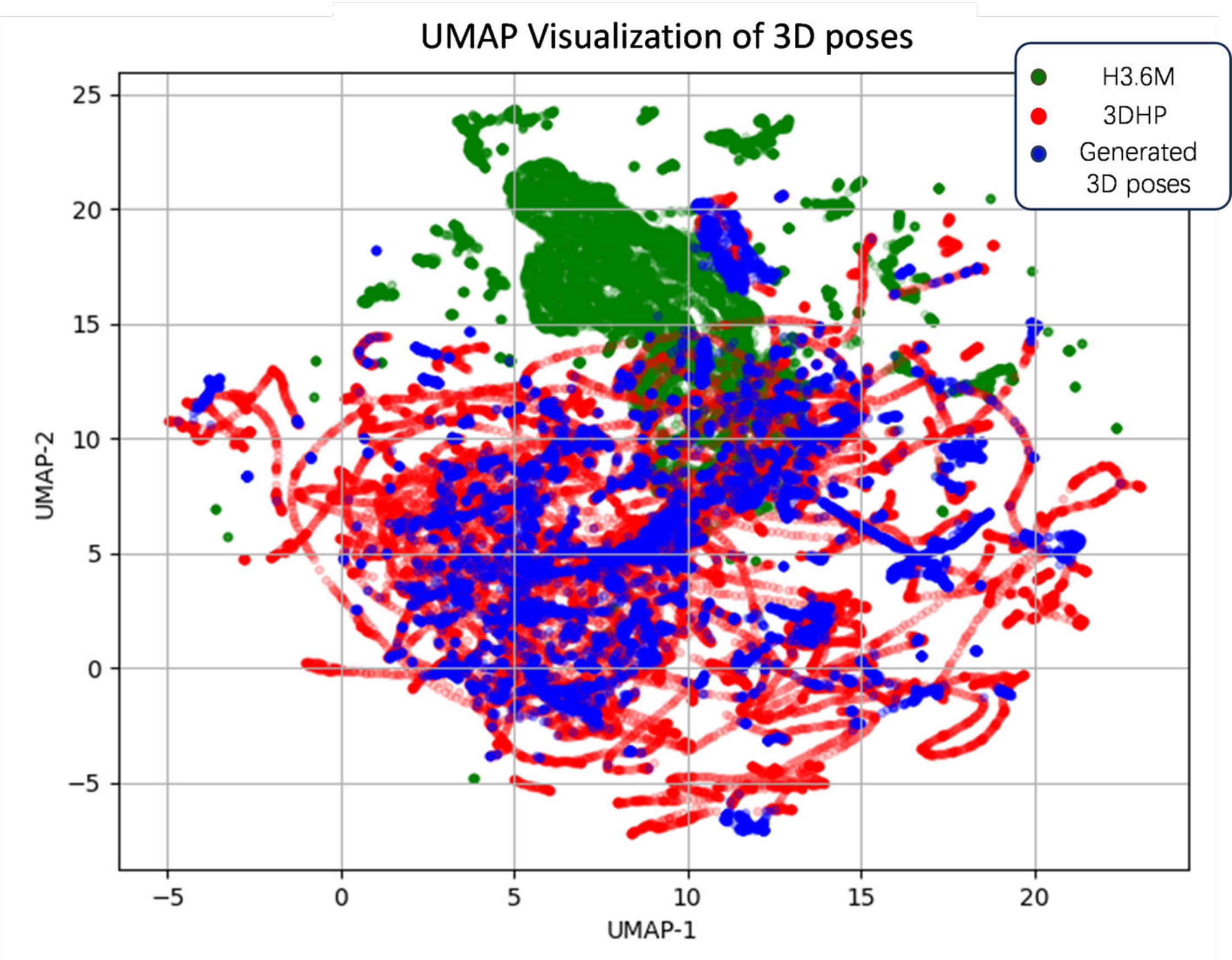}
   \caption{{\bf Distribution of 3D poses from H3.6M, 3DHP, and generated data.} The 2D embeddings are produced with UMAP \cite{umap}. Our generated data is more distributionally aligned with the target domain (3DHP), exhibiting greater overlap.}
   \label{fig-umap}
\end{figure}

Next, we use UMAP \cite{umap} to perform a non-linear embedding into 2D and visualize the data distribution in source H3.6M, target 3DHP, and the generated {3D poses}. As shown in Fig. \ref{fig-umap}, the generated {3D poses} shows higher overlap with 3DHP than with H3.6M, indicating significant effectiveness and consistency with the target domain.

\begin{figure}[!htp]
  \centering
   \includegraphics[width=8cm]{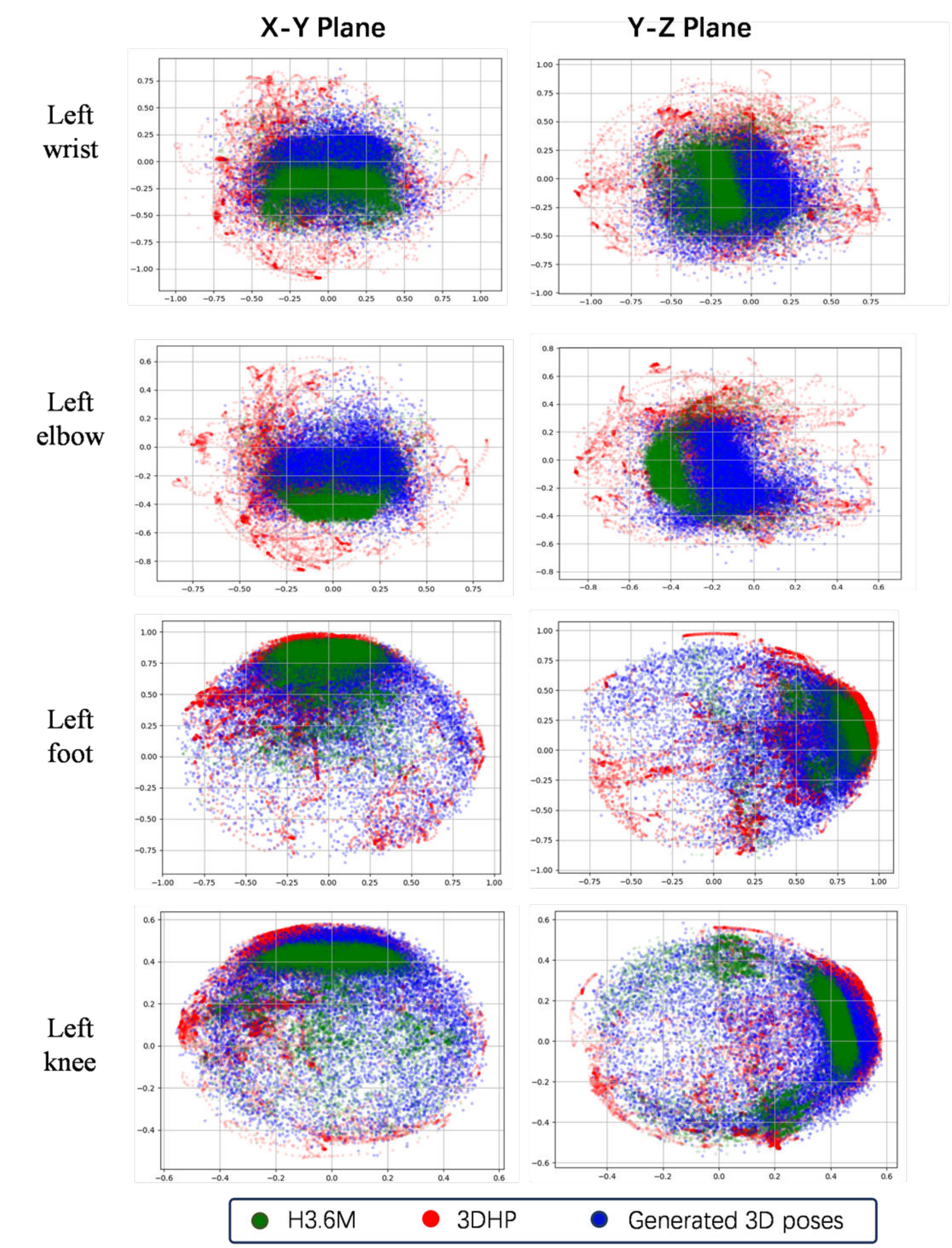}
   \caption{\bf{Distribution of the locations of joints from H3.6M, 3DHP, and generated {3D poses.}
   }  }
   \label{fig-joints}
\end{figure}

In addition, we provide a more intuitive visualization of the coordinate distributions of selected distal joints after root subtraction. As shown in Fig. \ref{fig-joints},  the joint location distribution of the generated 3D poses closely matches that of 3DHP, demonstrating its effectiveness in simulating the target domain.

\begin{figure}[!htp]
  \centering
   \includegraphics[width=8cm]{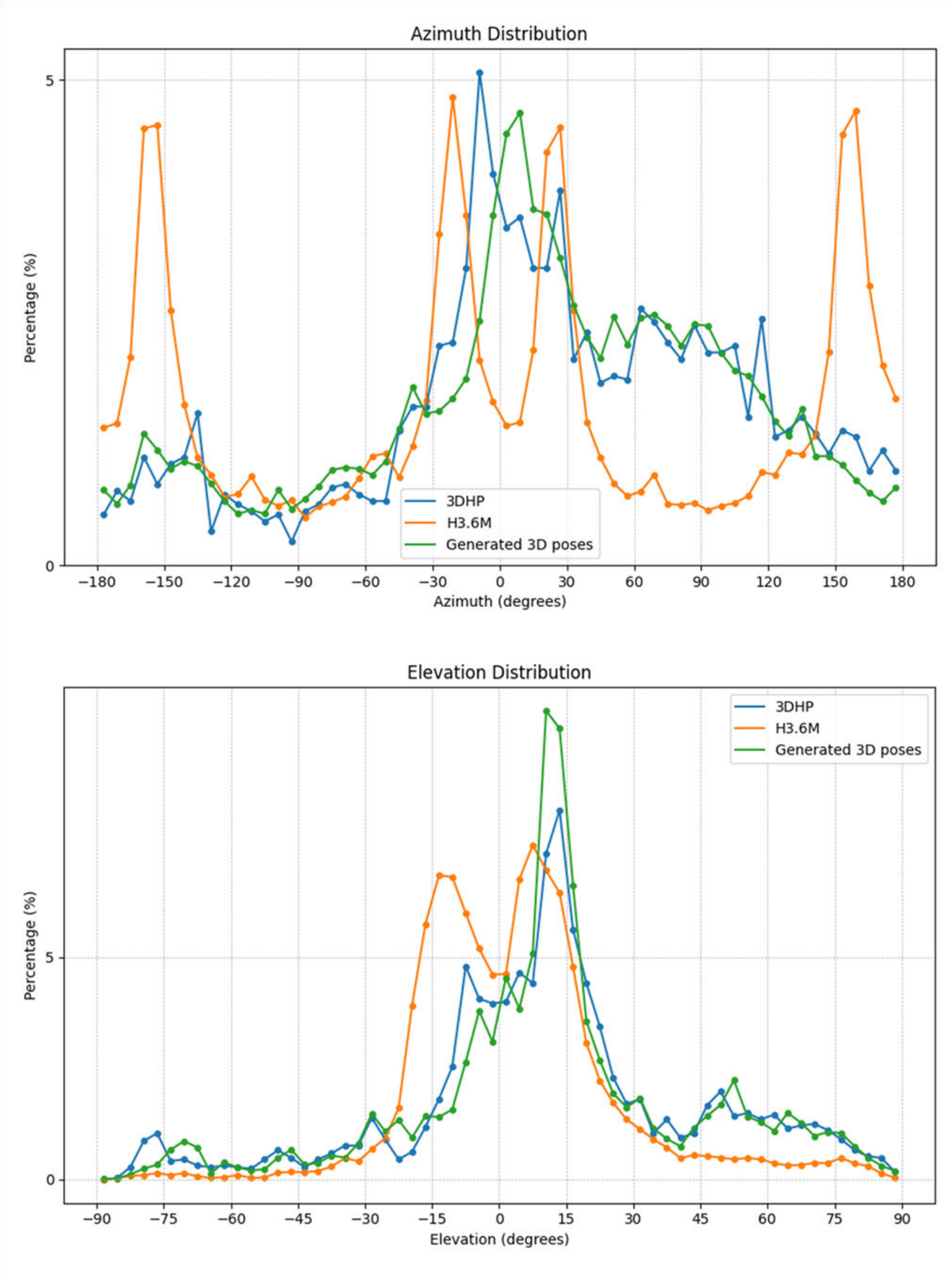}
   \caption{\bf{Distribution of camera viewpoint azimuth ($ -180^{\circ }, 180^{\circ }$) and elevation ($-90^{\circ }, 90^{\circ }$) from H3.6M, 3DHP and generated {3D poses.}
   }  }
   \label{fig-camera}
\end{figure}

Furthermore, as shown in Fig. \ref{fig-camera}, we visualize the discrepancy in camera viewpoints by computing the histogram of camera viewpoint azimuth and elevation relative to the body-centered coordinate frame. The {generated 3D poses mimic} the camera viewpoint distribution of the target domain, further supporting its effectiveness in bridging the domain gap.

\subsection{Ablation Study}
We conduct a series of ablation studies with H3.6M as the source domain and 3DHP as the target domain. We employ the ground truth 2D poses as input and VideoPose3D \cite{pavllo20193d} as the Pose Estimator.

\textbf{Effect of method's modules.}
We conduct an ablation study starting from a baseline model that was trained on the source domain, with our method's modules  systematically integrated to analyze their effects:
Pose Augmentor $M_a$, Pseudo-Label Generation module $M_p$, Global Transformation module $M_t$, and iterative training.
Since the generated 3D pseudo-labels approximate the target distribution and guide the global transformation, Pseudo-Label Generation module $M_p$ and Global Transformation module $M_t$ are evaluated as a unified component.

As shown in Table \ref{tab5}, when our Pose Augmentor $M_a$ is added we achieve an improvement of 6.1mm$\downarrow$ in MPJPE, followed by a substantial error reduction of 20.6mm$\downarrow$ when
the Global Transformation module $M_t$ with pseudo-labels from $M_p$ is introduced. Finally, through an iteration of the whole process, a further drop of {0.9mm} in MPJPE is achieved. Corresponding increases in PCK and AUC are also shown.

\begin{table}[!htp]
\setlength{\tabcolsep}{3.9pt}
\centering
\caption{{\bf The impact of each module in the proposed method.} The * indicates that the modules were processed for an extra iteration.}
\begin{tabular}{l|ccc}
\toprule
\bf{Method}  & \bf{MPJPE$\downarrow$} & \bf{PCK$\uparrow$} &\bf{AUC$\uparrow$} \\ \hline
VideoPose3D \cite{pavllo20193d}                                                           & 85.3 &84.1 & 52.9 \\
 VideoPose3D \cite{pavllo20193d} + {$M_a$}      & 79.2 &86.5& 54.9 \\
VideoPose3D \cite{pavllo20193d}  + {$M_a$} + ({$M_p$ \& $M_t$})                      & 58.6 &91.8 &64.3 \\
VideoPose3D \cite{pavllo20193d} + {$M_a$} +  ({$M_p$ \& $M_t$})*      & \bf{57.7} &\bf{92.5} &\bf{64.6} \\
 \bottomrule
\end{tabular}
\label{tab5}
\end{table}

\textbf{Different Global Transformations.} {The global transformation module $M_t$ contributes significantly to reducing the domain differences in camera positions and orientations from source to target. }
It is target-oriented and uses the pseudo-labels to guide the rotation and translation of the 3D poses. We compare three other approaches for global transformation: RR+RT, GR+RT, RR+GT, where RR denotes rotation matrices are randomly initialized and learned in the network, RT denotes translation matrices are randomly initialized and learned in the network, GR denotes the pseudo-label guided rotation and GT denotes the pseudo-label guided translation. Our approach is therefore GR+GT, which performs best as shown in Table \ref{tab7}.

\begin{table}[!htp]
\centering
\caption{\bf{The impact of different approaches for Global Transformation $M_t$.}}
\begin{tabular}{c|ccc}
\toprule
\bf{Global Transformation} & \bf{MPJPE$\downarrow$} & \bf{PCK$\uparrow$}  & \bf{AUC$\uparrow$}  \\ \hline
RR+RT                      & 73.0  & 88.6 & 57.3 \\
GR+RT                      & 66.0  & 89.9 & 60.9 \\
RR+GT                      & 63.7  & 90.3 & 62.0 \\
GR+GT                      & \bf{57.7}  & \bf{92.5} & \bf{64.6} \\
 \bottomrule
\end{tabular}
\label{tab7}
\end{table}

\textbf{Number of iterations.}
The influence of the number of iterations on the model's performance is shown in Table \ref{tab8}.
The results are averaged after five runs with different random seed values.
Beyond one round of iteration, the model fails to gain further increments and a possible reason could be overfitting.

\begin{table}[!htp]
\centering
\caption{\bf{The impact of using different number of iterations.}}
\begin{tabular}{lc|ccc}
\hline
\bf{Iterations} & \bf{MPJPE$\downarrow$ }& \bf{PCK$\uparrow$}  & \bf{AUC$\uparrow$ }  \\ \hline
0                   & 58.6  & 91.8 & 64.3 \\
1                   & \bf{57.7}  & \bf{92.5} & \bf{64.6} \\
2                   & 58.2  & 92.0 & 64.3 \\
3                   &58.0   & 92.3 & 64.4 \\
4                   & 58.3  & 92.0 & 64.3 \\
\bottomrule
\end{tabular}
\label{tab8}
\end{table}

{\textbf{Different hyparameters of the loss function.}
The impact of different hyparameter settings in the Generator loss function is tested and reported in Table \ref{tab9}. Based on our empirical investigation, we set $\alpha=1$, $\beta=6$ in the model. }

\begin{table}[!htp]
\centering
\caption{\bf{The impact of different hyparameters in Generator loss function.}}
\begin{tabular}{ll|lll}
\toprule
\bf{$\alpha$} & \bf{$\beta$} &\bf{MPJPE$\downarrow$ }& \bf{PCK$\uparrow$}  & \bf{AUC$\uparrow$ } \\ \hline
1     & 1/3  & 63.5  & 90.6 & 62.6 \\
1     & 1    & 62.4  & 91.0 & 62.7 \\
1     & 3    & 59.7  & 91.7 & 63.2 \\
\bf{1}     & \bf{6}    & \bf{57.7}  & \bf{92.5} & \bf{64.6} \\
1     & 9    & 58.2  & 92.3 & 64.4 \\
1     & 18   & 58.8  & 92.1 & 64.1 \\
\bottomrule
\end{tabular}
\label{tab9}
\end{table}

\subsection{Analysis of different Global Transformation solutions}
\label{Diff_GT}

We now analyse the reliability of our proposed cross-domain 3D pose alignment to achieve a global transformation against the less computationally efficient Kabsch algorithm \cite{kabsch1976solution, umeyama2002least}.

Given the 3D pose sets of the H3.6M dataset and the pseudo-3D pose sets of the 3DHP dataset, we randomly sample an equal number of poses from the H3.6M dataset to match the sample size of the 3DHP dataset. Based on these two equally sized 3D pose sets, we perform random pairwise alignment between them. This process is repeated five times to ensure robustness, and MPJPE is used to quantify the distance between the two sets.

As shown in Table \ref{analysis}, the MPJPE after global transformation alignment shows an average of 219.1mm which is only slightly higher than that of the Kabsch algorithm at 193.9mm.

\begin{table}[!htp]
\centering
\caption{{\bf MPJPE (mm) before and after random pairwise global transformation (GT) alignment performed five times.} The percentages show the improvement achieved against the `Before GT' value.}
\scalebox{0.85}{
\begin{tabular}{l|rrrrrr}
\hline
                         & {\bf 1} & {\bf 2} & {\bf 3} & {\bf 4} & {\bf 5} &{\bf Average $\downarrow$}  \\ \hline
Before GT                & 2096.6  & 2103.1  & 2117.1  & 2093.2  & 2098.2 & 2101.6    \\ \hline
\makecell[l]{GT using Kabsch \\method \cite{kabsch1976solution, umeyama2002least} } & 193.0  & 194.1  & 193.7  & 193.3  & 195.5 &
\makecell{193.9\\(90.8\%)}
\\ \hline
\makecell[l]{GT using \\ proposed method} & 218.1  & 219.4  & 218.7  &  218.6 &  220.8 & \makecell{ 219.1\\(89.6\%)}  \\ \hline
\end{tabular}
}
\label{analysis}
\end{table}

\subsection{Validation of the generated data}
To further validate the influence of the generated pseudo-labels on the model, we conducted experiments to train the pose estimator using varying amounts of both the generated pose data and the available target domain pose data. These experiments were carried out using VideoPose3D\cite{pavllo20193d} as the Pose Estimator, with H3.6M as the source domain and 3DHP as the target. The generated pose data was derived from the source domain, at a sample size that matches that of the original source dataset.

As shown in the light-grey part of Table \ref{generate_all}, for all different proportions of target data deployed, the performance of the model improved across all measures when the generated data was incorporated. In the darker-grey part of Table \ref{generate_all}, it can be observed that pose estimation performance again improves consistently across all measures when using all the target data and different proportions of generated pose data, ranging from 0\% to 100\%. For each trial, the proportion of generated data was randomly sampled.

\begin{table}[!htp]
\centering
\caption{\bf{Training the pose estimator using various percentages of generated pose data and target pose data.}}
\begin{tabular}{cc|r|r|r} \hline
{\bf Generated data} & {\bf Target data} &  {\bf MPJPE$\mathrm{\downarrow}$} & {\bf PCK$\mathrm{\uparrow}$} & {\bf AUC$\mathrm{\uparrow}$} \\ \hline
\rowcolor[gray]{.92} 0\% & 20\% &  82.5    & 86.0 & 52.7 \\ \hline
\rowcolor[gray]{.92} 100\% & 20\% &  \textbf{51.7}   & \textbf{93.9} & \textbf{67.6} \\ \hline \hline
\rowcolor[gray]{.92} 0\% & 40\% &  62.8      & 91.6 & 61.6 \\ \hline
\rowcolor[gray]{.92} 100\% & 40\% &  \textbf{48.4}    & \textbf{94.7} & \textbf{69.3} \\ \hline \hline
\rowcolor[gray]{.92}  0\% & 60\% & 54.8     & 93.7 & 65.7 \\ \hline
\rowcolor[gray]{.92} 100\% & 60\% &  \textbf{47.6} & \textbf{94.9} & \textbf{69.7} \\ \hline \hline
\rowcolor[gray]{.92} 0\% & 80\% &  50.6     & 94.6 & 67.9 \\ \hline
\rowcolor[gray]{.92}  100\% &  80\% &\textbf{45.8} & \textbf{95.3} & \textbf{70.7} \\ \hline
\rowcolor[gray]{.92}  0\% & 100\% & 49.0     & 94.9 & 68.9 \\ \hline
\rowcolor[gray]{.92} 100\% & 100\% &  \textbf{45.0}& \textbf{95.4} & \textbf{71.0} \\ \hline \hline
\rowcolor[gray]{.80} 0\% & 100\% &  49.0     & 94.9 & 68.9 \\ \hline
\rowcolor[gray]{.80} 20\% & 100\% &  47.2   & 95.2 & 69.7 \\ \hline
\rowcolor[gray]{.80} 40\% & 100\% &  46.4   & 95.3 & 70.1 \\ \hline
\rowcolor[gray]{.80} 60\% & 100\% & 45.8    & 95.4 & 70.5 \\ \hline
\rowcolor[gray]{.80} 80\% & 100\% &  45.3    & 95.4 & 70.9 \\ \hline
\rowcolor[gray]{.80}100\% & 100\% & \textbf{45.0}& \textbf{95.4} & \textbf{71.0} \\ \hline
\end{tabular}\
\label{generate_all}
\end{table}

\subsection{Discussion Compared to SOTA Methods}

The most recent state-of-the-art works in the field of unsupervised domain adaptation for 3D human pose estimation include PoseDA \cite{chai2023global}, PoSynDA \cite{liu2023posynda}, and the Dual-Augmentor framework \cite{peng2024dual}. Our method presents notable differences and advantages over them.

PoseDA \cite{chai2023global} aligns global positions of 3D poses to ensure their projected 2D poses match the target dataset in scale and position. While this aligns 2D poses across domains, it does not fully address the inherent ambiguity in mapping 2D to 3D. Our method aligns both 2D and 3D poses, ensuring a closer match between source and target domains in both input and label spaces.
PoSynDA\cite{liu2023posynda} builds on PoseDA by leveraging a diffusion-based model (D3DP \cite{shan2023diffusion}) to generate multiple pseudo-labels for the target domain. However, using a diffusion model increases model complexity, whereas, our approach introduces no additional training parameters in the Global Transformation module $M_t$.

The Dual-Augmentor framework in \cite{peng2024dual} generates diverse out-of-source distributions using a weak and strong augmentor,
while our approach fully utilizes target-domain 2D poses to generate pseudo-3D poses and {achieves} better cross-domain performance. {It is} primarily applicable when target data are accessible.

\section{Conclusion}
This paper proposes an iterative framework for domain-adaptive human pose estimation, focusing on aligning the 2D and 3D poses across domains to bridge domain gaps. Leveraging 2D poses from the target domain, we investigate a relatively underexplored strategy for estimating absolute 3D poses as {pseudo-labels.} These target-domain pseudo-labels provide effective guidance for the Global Transformation module, which incorporates a human-centered coordinate system to explicitly align global coordinates between the source and target domains.
This design is both intuitive and effective, achieving significant performance gains without adding extra training parameters. Additionally, we enhance pose augmentation by modifying bone lengths and angles, and we propose to use an adversarial learning architecture based on the target domain's 2D poses and the source domain's 3D poses.
The above modules are cohesively integrated into a unified framework that effectively enhances cross-domain pose estimation performance.

Extensive experiments on three public datasets demonstrate that our approach can surpass or compete with state-of-the-art models and outperform the target-domain trained model MixSTE\cite{zhang2022mixste}. This included experiments where we added a simulated,  challenging camera viewpoint to the 3DHP dataset to validate our method's robustness in even more difficult scenarios. For future work, we shall explore source-free domain adaptive human pose estimation without requiring access to source data.

\section*{Acknowledgments}
This work was supported by the TORUS Project, which has been funded by the UK Engineering and Physical Sciences Research Council (EPSRC), grant number EP/X036146/1.

\ifCLASSOPTIONcaptionsoff
  \newpage
\fi


\begin{thebibliography}{00}
\normalem
\bibitem{jiao2020pen}
Y.~Jiao, H.~Yao, and C.~Xu, ``Pen: Pose-embedding network for pedestrian
  detection,'' \emph{IEEE Transactions on Circuits and Systems for Video
  Technology}, vol.~31, no.~3, pp. 1150--1162, 2020.

\bibitem{wu2023video}
L.~Wu, C.~Huang, L.~Fei, S.~Zhao, J.~Zhao, Z.~Cui, and Y.~Xu, ``Video-based
  fall detection using human pose and constrained generative adversarial
  network,'' \emph{IEEE Transactions on Circuits and Systems for Video
  Technology}, 2023.

\bibitem{guzov2021human}
V.~Guzov, A.~Mir, T.~Sattler, and G.~Pons-Moll, ``Human poseitioning system
  (hps): 3d human pose estimation and self-localization in large scenes from
  body-mounted sensors,'' in \emph{Proceedings of the IEEE/CVF Conference on
  Computer Vision and Pattern Recognition}, 2021, pp. 4318--4329.

\bibitem{yi2023mime}
H.~Yi, C.-H.~P. Huang, S.~Tripathi, L.~Hering, J.~Thies, and M.~J. Black,
  ``Mime: Human-aware 3d scene generation,'' in \emph{Proceedings of the
  IEEE/CVF Conference on Computer Vision and Pattern Recognition}, 2023, pp.
  12\,965--12\,976.

\bibitem{haouchine2021pose}
N.~Haouchine, P.~Juvekar, M.~Nercessian, W.~M. Wells~III, A.~Golby, and
  S.~Frisken, ``Pose estimation and non-rigid registration for augmented
  reality during neurosurgery,'' \emph{IEEE Transactions on Biomedical
  Engineering}, vol.~69, no.~4, pp. 1310--1317, 2021.

\bibitem{yu2023toward}
H.~Yu, X.~Fan, Y.~Hou, W.~Pei, H.~Ge, X.~Yang, D.~Zhou, Q.~Zhang, and M.~Zhang,
  ``Toward realistic 3d human motion prediction with a spatio-temporal
  cross-transformer approach,'' \emph{IEEE Transactions on Circuits and Systems
  for Video Technology}, vol.~33, no.~10, pp. 5707--5720, 2023.

\bibitem{lu2023hard}
Z.~Lu, H.~Wang, Z.~Chang, G.~Yang, and H.~P. Shum, ``Hard no-box adversarial
  attack on skeleton-based human action recognition with
  skeleton-motion-informed gradient,'' in \emph{Proceedings of the IEEE/CVF
  International Conference on Computer Vision}, 2023, pp. 4597--4606.

\bibitem{zimmermann20183d}
C.~Zimmermann, T.~Welschehold, C.~Dornhege, W.~Burgard, and T.~Brox, ``3d human
  pose estimation in rgbd images for robotic task learning,'' in \emph{2018
  IEEE International Conference on Robotics and Automation (ICRA)}.\hskip 1em
  plus 0.5em minus 0.4em\relax IEEE, 2018, pp. 1986--1992.

\bibitem{liu2021collision}
H.~Liu and L.~Wang, ``Collision-free human-robot collaboration based on context
  awareness,'' \emph{Robotics and Computer-Integrated Manufacturing}, vol.~67,
  p. 101997, 2021.

\bibitem{svenstrup2009pose}
M.~Svenstrup, S.~Tranberg, H.~J. Andersen, and T.~Bak, ``Pose estimation and
  adaptive robot behaviour for human-robot interaction,'' in \emph{2009 IEEE
  International Conference on Robotics and Automation}.\hskip 1em plus 0.5em
  minus 0.4em\relax IEEE, 2009, pp. 3571--3576.

\bibitem{huo20233d}
R.~Huo, Q.~Gao, J.~Qi, and Z.~Ju, ``3d human pose estimation in video for
  human-computer/robot interaction,'' in \emph{International Conference on
  Intelligent Robotics and Applications}.\hskip 1em plus 0.5em minus
  0.4em\relax Springer, 2023, pp. 176--187.

\bibitem{gu2019multi}
R.~Gu, G.~Wang, Z.~Jiang, and J.-N. Hwang, ``Multi-person hierarchical 3d pose
  estimation in natural videos,'' \emph{IEEE Transactions on Circuits and
  Systems for Video Technology}, vol.~30, no.~11, pp. 4245--4257, 2019.

\bibitem{wandt2021canonpose}
B.~Wandt, M.~Rudolph, P.~Zell, H.~Rhodin, and B.~Rosenhahn, ``Canonpose:
  Self-supervised monocular 3d human pose estimation in the wild,'' in
  \emph{Proceedings of the IEEE/CVF conference on computer vision and pattern
  recognition}, 2021, pp. 13\,294--13\,304.

\bibitem{tang2023ftcm}
Z.~Tang, Y.~Hao, J.~Li, and R.~Hong, ``Ftcm: Frequency-temporal collaborative
  module for efficient 3d human pose estimation in video,'' \emph{IEEE
  Transactions on Circuits and Systems for Video Technology}, vol.~34, no.~2,
  pp. 911--923, 2023.

\bibitem{zhou2023dual}
L.~Zhou, Y.~Chen, and J.~Wang, ``Dual-path transformer for 3d human pose
  estimation,'' \emph{IEEE Transactions on Circuits and Systems for Video
  Technology}, 2023.

\bibitem{zhang2020inference}
J.~Zhang, X.~Nie, and J.~Feng, ``Inference stage optimization for
  cross-scenario 3d human pose estimation,'' \emph{Advances in neural
  information processing systems}, vol.~33, pp. 2408--2419, 2020.

\bibitem{gong2021poseaug}
K.~Gong, J.~Zhang, and J.~Feng, ``Poseaug: A differentiable pose augmentation
  framework for 3d human pose estimation,'' in \emph{Proceedings of the
  IEEE/CVF conference on computer vision and pattern recognition}, 2021, pp.
  8575--8584.

\bibitem{gholami2022adaptpose}
M.~Gholami, B.~Wandt, H.~Rhodin, R.~Ward, and Z.~J. Wang, ``Adaptpose:
  Cross-dataset adaptation for 3d human pose estimation by learnable motion
  generation,'' in \emph{Proceedings of the IEEE/CVF Conference on Computer
  Vision and Pattern Recognition}, 2022, pp. 13\,075--13\,085.

\bibitem{peng2023source}
Q.~Peng, C.~Zheng, and C.~Chen, ``Source-free domain adaptive human pose
  estimation,'' in \emph{Proceedings of the IEEE/CVF International Conference
  on Computer Vision}, 2023, pp. 4826--4836.

\bibitem{peng2024dual}
------, ``A dual-augmentor framework for domain generalization in 3d human pose
  estimation,'' in \emph{Proceedings of the IEEE/CVF Conference on Computer
  Vision and Pattern Recognition}, 2024, pp. 2240--2249.

\bibitem{chai2023global}
W.~Chai, Z.~Jiang, J.-N. Hwang, and G.~Wang, ``Global adaptation meets local
  generalization: Unsupervised domain adaptation for 3d human pose
  estimation,'' in \emph{Proceedings of the IEEE/CVF International Conference
  on Computer Vision}, 2023, pp. 14\,655--14\,665.

\bibitem{liu2023posynda}
H.~Liu, J.-Y. He, Z.-Q. Cheng, W.~Xiang, Q.~Yang, W.~Chai, G.~Wang, X.~Bao,
  B.~Luo, Y.~Geng \emph{et~al.}, ``Posynda: Multi-hypothesis pose synthesis
  domain adaptation for robust 3d human pose estimation,'' in \emph{Proceedings
  of the 31st ACM International Conference on Multimedia}, 2023, pp.
  5542--5551.

\bibitem{pavllo20193d}
D.~Pavllo, C.~Feichtenhofer, D.~Grangier, and M.~Auli, ``3d human pose
  estimation in video with temporal convolutions and semi-supervised
  training,'' in \emph{Proceedings of the IEEE/CVF conference on computer
  vision and pattern recognition}, 2019, pp. 7753--7762.

\bibitem{zhang2022mixste}
J.~Zhang, Z.~Tu, J.~Yang, Y.~Chen, and J.~Yuan, ``Mixste: Seq2seq mixed
  spatio-temporal encoder for 3d human pose estimation in video,'' in
  \emph{Proceedings of the IEEE/CVF conference on computer vision and pattern
  recognition}, 2022, pp. 13\,232--13\,242.

\bibitem{lin2021end}
K.~Lin, L.~Wang, and Z.~Liu, ``End-to-end human pose and mesh reconstruction
  with transformers,'' in \emph{Proceedings of the IEEE/CVF conference on
  computer vision and pattern recognition}, 2021, pp. 1954--1963.

\bibitem{jin2022single}
L.~Jin, C.~Xu, X.~Wang, Y.~Xiao, Y.~Guo, X.~Nie, and J.~Zhao, ``Single-stage is
  enough: Multi-person absolute 3d pose estimation,'' in \emph{Proceedings of
  the IEEE/CVF Conference on Computer Vision and Pattern Recognition}, 2022,
  pp. 13\,086--13\,095.

\bibitem{li2023niki}
J.~Li, S.~Bian, Q.~Liu, J.~Tang, F.~Wang, and C.~Lu, ``Niki: Neural inverse
  kinematics with invertible neural networks for 3d human pose and shape
  estimation,'' in \emph{Proceedings of the IEEE/CVF Conference on Computer
  Vision and Pattern Recognition}, 2023, pp. 12\,933--12\,942.

\bibitem{sarandi2023learning}
I.~S{\'a}r{\'a}ndi, A.~Hermans, and B.~Leibe, ``Learning 3d human pose
  estimation from dozens of datasets using a geometry-aware autoencoder to
  bridge between skeleton formats,'' in \emph{Proceedings of the IEEE/CVF
  Winter Conference on Applications of Computer Vision}, 2023, pp. 2956--2966.

\bibitem{zheng20213d}
C.~Zheng, S.~Zhu, M.~Mendieta, T.~Yang, C.~Chen, and Z.~Ding, ``3d human pose
  estimation with spatial and temporal transformers,'' in \emph{Proceedings of
  the IEEE/CVF international conference on computer vision}, 2021, pp.
  11\,656--11\,665.

\bibitem{zou2021modulated}
Z.~Zou and W.~Tang, ``Modulated graph convolutional network for 3d human pose
  estimation,'' in \emph{Proceedings of the IEEE/CVF international conference
  on computer vision}, 2021, pp. 11\,477--11\,487.

\bibitem{li2022mhformer}
W.~Li, H.~Liu, H.~Tang, P.~Wang, and L.~Van~Gool, ``Mhformer: Multi-hypothesis
  transformer for 3d human pose estimation,'' in \emph{Proceedings of the
  IEEE/CVF Conference on Computer Vision and Pattern Recognition}, 2022, pp.
  13\,147--13\,156.

\bibitem{tang20233d}
Z.~Tang, Z.~Qiu, Y.~Hao, R.~Hong, and T.~Yao, ``3d human pose estimation with
  spatio-temporal criss-cross attention,'' in \emph{Proceedings of the IEEE/CVF
  Conference on Computer Vision and Pattern Recognition}, 2023, pp. 4790--4799.

\bibitem{sun2019deep}
K.~Sun, B.~Xiao, D.~Liu, and J.~Wang, ``Deep high-resolution representation
  learning for human pose estimation,'' in \emph{Proceedings of the IEEE/CVF
  conference on computer vision and pattern recognition}, 2019, pp. 5693--5703.

\bibitem{li2020cascaded}
S.~Li, L.~Ke, K.~Pratama, Y.-W. Tai, C.-K. Tang, and K.-T. Cheng, ``Cascaded
  deep monocular 3d human pose estimation with evolutionary training data,'' in
  \emph{Proceedings of the IEEE/CVF conference on computer vision and pattern
  recognition}, 2020, pp. 6173--6183.

\bibitem{jin2020whole}
S.~Jin, L.~Xu, J.~Xu, C.~Wang, W.~Liu, C.~Qian, W.~Ouyang, and P.~Luo,
  ``Whole-body human pose estimation in the wild,'' in \emph{Computer
  Vision--ECCV 2020: 16th European Conference, Glasgow, UK, August 23--28,
  2020, Proceedings, Part IX 16}.\hskip 1em plus 0.5em minus 0.4em\relax
  Springer, 2020, pp. 196--214.

\bibitem{martinez2017simple}
J.~Martinez, R.~Hossain, J.~Romero, and J.~J. Little, ``A simple yet effective
  baseline for 3d human pose estimation,'' in \emph{Proceedings of the IEEE
  international conference on computer vision}, 2017, pp. 2640--2649.

\bibitem{fang2022alphapose}
H.-S. Fang, J.~Li, H.~Tang, C.~Xu, H.~Zhu, Y.~Xiu, Y.-L. Li, and C.~Lu,
  ``Alphapose: Whole-body regional multi-person pose estimation and tracking in
  real-time,'' \emph{IEEE Transactions on Pattern Analysis and Machine
  Intelligence}, vol.~45, no.~6, pp. 7157--7173, 2022.

\bibitem{lu2024rtmo}
P.~Lu, T.~Jiang, Y.~Li, X.~Li, K.~Chen, and W.~Yang, ``Rtmo: Towards
  high-performance one-stage real-time multi-person pose estimation,'' in
  \emph{Proceedings of the IEEE/CVF Conference on Computer Vision and Pattern
  Recognition}, 2024, pp. 1491--1500.

\bibitem{liu2024human}
S.~Liu, X.~Xie, and G.~Shi, ``Human pose estimation via parse graph of body
  structure,'' \emph{IEEE Transactions on Circuits and Systems for Video
  Technology}, 2024.

\bibitem{shan2023diffusion}
W.~Shan, Z.~Liu, X.~Zhang, Z.~Wang, K.~Han, S.~Wang, S.~Ma, and W.~Gao,
  ``Diffusion-based 3d human pose estimation with multi-hypothesis
  aggregation,'' in \emph{Proceedings of the IEEE/CVF International Conference
  on Computer Vision}, 2023, pp. 14\,761--14\,771.

\bibitem{cai2024}
Q.~Cai, X.~Hu, S.~Hou, Y.~Li, and Y.~Huang, ``Disentangled diffusion-based 3d
  human pose estimation with hierarchical spatial and temporal denoiser,'' in
  \emph{Proceedings of the AAAI Conference on Artificial Intelligence}, 2024,
  pp. 882--890.

\bibitem{rogez2016mocap}
G.~Rogez and C.~Schmid, ``Mocap-guided data augmentation for 3d pose estimation
  in the wild,'' \emph{Advances in neural information processing systems},
  vol.~29, 2016.

\bibitem{mehta2017vnect}
D.~Mehta, S.~Sridhar, O.~Sotnychenko, H.~Rhodin, M.~Shafiei, H.-P. Seidel,
  W.~Xu, D.~Casas, and C.~Theobalt, ``Vnect: Real-time 3d human pose estimation
  with a single rgb camera,'' \emph{Acm transactions on graphics (tog)},
  vol.~36, no.~4, pp. 1--14, 2017.

\bibitem{xu2021monocular}
Y.~Xu, W.~Wang, T.~Liu, X.~Liu, J.~Xie, and S.-C. Zhu, ``Monocular 3d pose
  estimation via pose grammar and data augmentation,'' \emph{IEEE transactions
  on pattern analysis and machine intelligence}, vol.~44, no.~10, pp.
  6327--6344, 2021.

\bibitem{yang2018body}
B.~Yang, A.~J. Ma, and P.~C. Yuen, ``Body parts synthesis for cross-quality
  pose estimation,'' \emph{IEEE Transactions on Circuits and Systems for Video
  Technology}, vol.~29, no.~2, pp. 461--474, 2018.

\bibitem{yang2023camerapose}
C.-Y. Yang, J.~Luo, L.~Xia, Y.~Sun, N.~Qiao, K.~Zhang, Z.~Jiang, J.-N. Hwang,
  and C.-H. Kuo, ``Camerapose: Weakly-supervised monocular 3d human pose
  estimation by leveraging in-the-wild 2d annotations,'' in \emph{Proceedings
  of the IEEE/CVF Winter Conference on Applications of Computer Vision}, 2023,
  pp. 2924--2933.

\bibitem{du2024joypose}
S.~Du, Z.~Yuan, P.~Lai, and T.~Ikenaga, ``Joypose: Jointly learning
  evolutionary data augmentation and anatomy-aware global--local representation
  for 3d human pose estimation,'' \emph{Pattern Recognition}, vol. 147, p.
  110116, 2024.

\bibitem{kimtoward}
J.-H. Kim and S.-W. Lee, ``Toward approaches to scalability in 3d human pose
  estimation,'' in \emph{The Thirty-eighth Annual Conference on Neural
  Information Processing Systems}, 2024.

\bibitem{guan2021bilevel}
S.~Guan, J.~Xu, Y.~Wang, B.~Ni, and X.~Yang, ``Bilevel online adaptation for
  out-of-domain human mesh reconstruction,'' in \emph{Proceedings of the
  IEEE/CVF Conference on Computer Vision and Pattern Recognition}, 2021, pp.
  10\,472--10\,481.

\bibitem{kabsch1976solution}
W.~Kabsch, ``A solution for the best rotation to relate two sets of vectors,''
  \emph{Foundations of Crystallography}, vol.~32, no.~5, pp. 922--923, 1976.

\bibitem{umeyama2002least}
S.~Umeyama, ``Least-squares estimation of transformation parameters between two
  point patterns,'' \emph{IEEE Transactions on pattern analysis and machine
  intelligence}, vol.~13, no.~4, pp. 376--380, 1991.

\bibitem{wang2020predicting}
Z.~Wang, D.~Shin, and C.~C. Fowlkes, ``Predicting camera viewpoint improves
  cross-dataset generalization for 3d human pose estimation,'' in
  \emph{Computer vision--ECCV 2020 workshops: Glasgow, UK, August 23--28, 2020,
  proceedings, part II 16}.\hskip 1em plus 0.5em minus 0.4em\relax Springer,
  2020, pp. 523--540.

\bibitem{wei2019view}
G.~Wei, C.~Lan, W.~Zeng, and Z.~Chen, ``View invariant 3d human pose
  estimation,'' \emph{IEEE Transactions on Circuits and Systems for Video
  Technology}, vol.~30, no.~12, pp. 4601--4610, 2020.

\bibitem{wandt2018kinematic}
B.~Wandt, H.~Ackermann, and B.~Rosenhahn, ``A kinematic chain space for
  monocular motion capture,'' in \emph{Proceedings of the European Conference
  on Computer Vision (ECCV) Workshops}, 2018, pp. 0--0.

\bibitem{wandt2019repnet}
B.~Wandt and B.~Rosenhahn, ``Repnet: Weakly supervised training of an
  adversarial reprojection network for 3d human pose estimation,'' in
  \emph{Proceedings of the IEEE/CVF conference on computer vision and pattern
  recognition}, 2019, pp. 7782--7791.

\bibitem{mao2017least}
X.~Mao, Q.~Li, H.~Xie, R.~Y. Lau, Z.~Wang, and S.~Paul~Smolley, ``Least squares
  generative adversarial networks,'' in \emph{Proceedings of the IEEE
  international conference on computer vision}, 2017, pp. 2794--2802.

\bibitem{von2018recovering}
T.~Von~Marcard, R.~Henschel, M.~J. Black, B.~Rosenhahn, and G.~Pons-Moll,
  ``Recovering accurate 3d human pose in the wild using imus and a moving
  camera,'' in \emph{Proceedings of the European conference on computer vision
  (ECCV)}, 2018, pp. 601--617.

\bibitem{ionescu2013human3}
C.~Ionescu, D.~Papava, V.~Olaru, and C.~Sminchisescu, ``Human3. 6m: Large scale
  datasets and predictive methods for 3d human sensing in natural
  environments,'' \emph{IEEE transactions on pattern analysis and machine
  intelligence}, vol.~36, no.~7, pp. 1325--1339, 2013.

\bibitem{kolotouros2019learning}
N.~Kolotouros, G.~Pavlakos, M.~J. Black, and K.~Daniilidis, ``Learning to
  reconstruct 3d human pose and shape via model-fitting in the loop,'' in
  \emph{Proceedings of the IEEE/CVF international conference on computer
  vision}, 2019, pp. 2252--2261.

\bibitem{kocabas2020vibe}
M.~Kocabas, N.~Athanasiou, and M.~J. Black, ``Vibe: Video inference for human
  body pose and shape estimation,'' in \emph{Proceedings of the IEEE/CVF
  conference on computer vision and pattern recognition}, 2020, pp. 5253--5263.

\bibitem{lin2021mesh}
K.~Lin, L.~Wang, and Z.~Liu, ``Mesh graphormer,'' in \emph{Proceedings of the
  IEEE/CVF international conference on computer vision}, 2021, pp.
  12\,939--12\,948.

\bibitem{umap}
L.~McInnes, J.~Healy, and J.~Melville, ``Umap: Uniform manifold approximation
  and projection for dimension reduction,'' \emph{arXiv preprint
  arXiv:1802.03426}, 2018.

\end{thebibliography}

\newpage

\section*{Biography Section}
\begin{IEEEbiographynophoto}{Jingjing Liu}
received the Ph.D degree from the School of Mechanical Engineering, Shanghai Jiao Tong University, Shanghai, China, in 2023. She is currently a research associate in University of Bristol, the United Kingdom. Her current research interests include computer vision, human activity recognition, pose estimation and applications in parkinson's monitoring.
\end{IEEEbiographynophoto}

\begin{IEEEbiographynophoto}
{Zhiyong Wang} (Member, IEEE) received the Ph.D. degree in mechanical engineering from the School of Mechanical Engineering, Shanghai Jiao Tong University, Shanghai, China, in 2023. He is currently an Assistant Professor with the Harbin Institute of Technology Shenzhen, Shenzhen, China. His research interests include human-computer interaction, brain functional response, gaze estimation, and their practical applications in medical rehabilitation.
\end{IEEEbiographynophoto}

\begin{IEEEbiographynophoto}{Xinyu Fan} is currently working toward the undergraduate degree in automation from Xiamen University, Fujian, China. His research interests include computer vision, deep learning, human pose estimation and 3D reconstruction.
\end{IEEEbiographynophoto}

\begin{IEEEbiographynophoto}{Amirhossein Dadashzadeh} is a Research Associate in Computer Vision at the University of Bristol. He holds a Ph.D. in Computer Science from the same institution. His research focuses on developing advanced algorithms for monitoring Parkinson’s patients by analysing human actions in video data.
\end{IEEEbiographynophoto}

\begin{IEEEbiographynophoto}{Honghai Liu} (Fellow, IEEE) received the Ph.D. degree in robotics from King’s College London, London, U.K., in 2003.
He is currently a Professor with the School of Mechanical Engineering and Automation, Harbin Institute of Technology Shenzhen, Shenzhen, China, and with the School of Computing, Portsmouth, U.K. His research interests include human-machine interaction, computer vision, machine learning and their clinical applications with an emphasis on approaches that could make contribution to the intelligent connection of perception to action using contextual information.
Prof. Liu was a member of European Academy of Sciences. He is an Associate Editor of IEEE TRANSACTIONS ON INDUSTRIAL INFORMATICS and the IEEE TRANSACTIONS ON CYBERNETICS.
\end{IEEEbiographynophoto}

\begin{IEEEbiographynophoto}{Majid Mirmehdi} is a Professor of Computer Vision in the School of Computer Science at the University of Bristol. His research interests include natural scene analysis, healthcare monitoring using vision and other sensors, and human and animal motion and behaviour understanding. He has more than 250 refereed journal and conference publications in these and other areas. MM is a Fellow of the International Association for Pattern Recognition and a Distinguished Fellow of the British Machine Vision Association. He is Editor-in-Chief of IET Computer Vision journal and an Associate Editor of Pattern Recognition journal. He is a member of the IET and serves on the Executive Committee of the British Machine Vision Association.
\end{IEEEbiographynophoto}

\vfill

\end{document}